\documentclass{article}

\usepackage[dandb, final]{neurips_2025}
\makeatletter
\providecommand{\@trackname}{}
\makeatother
\usepackage[utf8]{inputenc} % allow utf-8 input
\usepackage[T1]{fontenc}    % use 8-bit T1 fonts
\usepackage{url}            % simple URL typesetting
\usepackage{booktabs}       % professional-quality tables
\usepackage{amsfonts}       % blackboard math symbols
\usepackage{amsmath}
\usepackage{nicefrac}       % compact symbols for 1/2, etc.
\usepackage{microtype}      % microtypography
\usepackage{colortbl,xcolor}         % colors

\usepackage{caption} % Do not use hypcap=true option globally

\definecolor{customblue}{rgb}{0.25, 0.41, 0.88} %
\usepackage[pagebackref,breaklinks,colorlinks,allcolors=customblue,hypertexnames=false]{hyperref}

\usepackage[capitalise]{cleveref} % for \cref 
\usepackage{makecell}      % table cells formatting
\usepackage{enumitem} % bullet-point paragraph planning

\usepackage{graphicx}
\usepackage{tabularx}
\usepackage{multirow}
\usepackage{xspace}
\usepackage{booktabs}
\usepackage{array}
\usepackage{pifont}
\usepackage{makecell}
\usepackage{siunitx} % format values with unit

\DeclareSIUnit{\K}{K}
\DeclareSIUnit{\M}{M}
\DeclareSIUnit{\B}{B}

\crefname  {figure}{Fig.}{Figs.}
\crefname  {table} {Tab.}{Tabs.} 

\newcolumntype{C}[1]{>{\centering\arraybackslash}m{#1}} % vertical centering
\definecolor{tblgray}{gray}{0.92}

\definecolor{my_green}{RGB}{51,102,0}
\definecolor{my_red}{RGB}{204, 0, 0}
\renewcommand{\checkmark}{\textcolor{my_green}{\ding{51}}} % ✔
\newcommand{\crossmark}{\textcolor{my_red}{\ding{55}}} % ✘

\newcommand{\dataindoor}{SceneSplat-7K\xspace}
\newcommand{\dataall}{SceneSplat-49K\xspace}
\newcommand{\ourdata}{SceneSplat-49K\xspace}

\newcommand{\benchmark}{SceneSplat-Bench\xspace}

\def\scannet{ScanNet\xspace}
\def\scannetpp{ScanNet++\xspace}
\def\matt{Matterport3D\xspace}

\def\holi{HoliCity\xspace}

\newcommand{\boldparagraph}[1]{\vspace{0.1em}\noindent{\bf #1}}

\newcommand\blfootnote[1]{%
  \begingroup
  \renewcommand\thefootnote{}\footnote{#1}%
  \addtocounter{footnote}{-1}%
  \endgroup
}

\title{SceneSplat++: A Large Dataset and Comprehensive \\ Benchmark for Language Gaussian Splatting}

\author{
\small{
\footnotemark[1]~~Mengjiao Ma$^{1,2}$,
\footnotemark[1]~~Qi Ma$^{1,3}$,
\footnotemark[1]~~Yue Li$^{4}$,
Jiahuan Cheng$^{5}$, 
Runyi Yang$^{1}$,
\footnotemark[2]~~Bin Ren$^{1,6,7}$,
Nikola Popovic$^{1}$} \\
\small{
\textbf{Mingqiang Wei}$^{2}$, 
\textbf{Nicu Sebe}$^{7}$,
\textbf{Luc Van Gool}$^{1}$,
\textbf{Theo Gevers}$^{4}$, 
\textbf{Martin R. Oswald}$^{4}$, 
\textbf{Danda Pani Paudel}$^{1}$} \\
\scriptsize{
$^1$INSAIT, Sofia University “St. Kliment Ohridski”\;
$^2$Nanjing University of Aeronautics and Astronautics\;
$^3$ETH Z\"urich\;
} \\
\scriptsize{
$^4$University of Amsterdam\;
$^5$Johns Hopkins University\;
$^6$University of Pisa\;
$^7$University of Trento}\\
}
\begin{document}

\maketitle
\begin{center}
    \vspace{-1cm}
\includegraphics[width=1.0\linewidth]{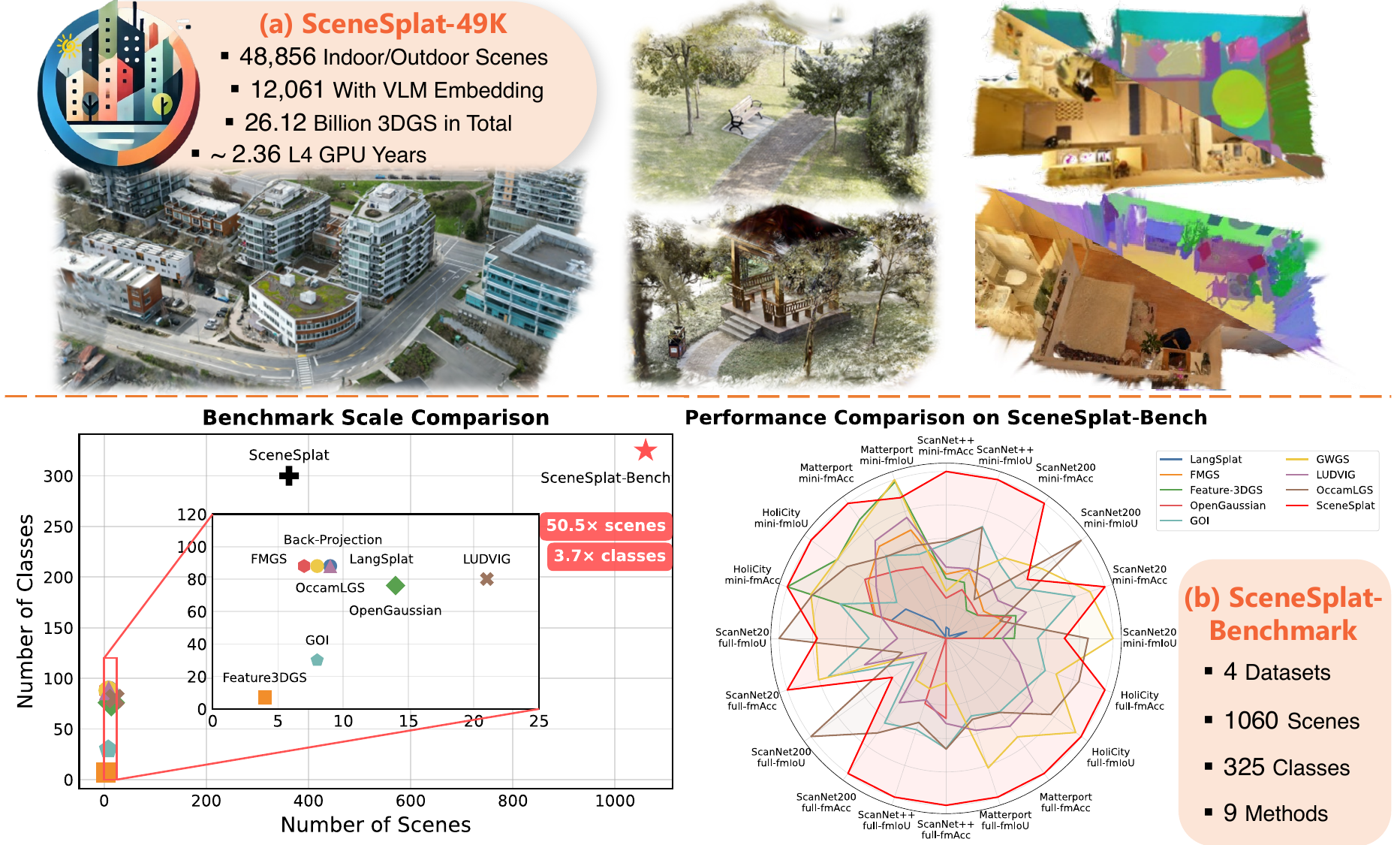}
     \captionof{figure}{\textbf{(a) \dataall Dataset}. We introduce \ourdata, a large-scale 3DGS dataset comprising approximately \qty{49}{\K} diverse indoor and outdoor scenes. {\textbf{(b) \benchmark}}. We introduce a comprehensive benchmark for evaluating \textit{Language Gaussian Splatting} (LGS) methods at scale. \textbf{Benchmark Scale Comparison} shows \benchmark significantly exceeds existing evaluation protocols with 3.7$\times$ more semantic classes and 50.5$\times$ more scenes. \textbf{Performance Comparison on \benchmark} using 4 datasets across 9 methods demonstrates that the generalizable approach(SceneSplat) consistently outperforms the per-scene methods in most of the benchmark variants. Our codes, benchmark, and datasets are released at \href{https://scenesplatpp.gaussianworld.ai/}{https://scenesplatpp.gaussianworld.ai/}.}
     \vspace{-2mm}
\label{fig:teaser}
\end{center}

\blfootnote{%
$\ast$ indicates equal contribution. %
$\dagger$ indicates the corresponding author: Bin Ren %
\textless\href{mailto:bin.ren@insait.ai}{bin.ren@insait.ai}\textgreater.%
}

% * Title: Is per-scene optimization necessary for Gaussian language splatting?

\begin{abstract}
%Establishing a connection between language and semantics provides the most effective approach to understanding 3D scenes.
\label{sec:abstract}
3D Gaussian Splatting (3DGS) serves as a highly performant and efficient encoding of scene geometry, appearance, and semantics. Moreover, grounding language in 3D scenes has proven to be an effective strategy for 3D scene understanding. Current \emph{Language Gaussian Splatting} line of work falls into three main groups: (i) per-scene optimization-based, (ii) per-scene optimization-free, and (iii) generalizable approach. However, most of them are evaluated only on rendered 2D views of a handful of scenes and viewpoints close to the training views, limiting their ability and insight into holistic 3D understanding. To address this gap, we propose the first large-scale benchmark that systematically assesses these three groups of methods directly in 3D space, evaluating on 1060 scenes across three indoor datasets and one outdoor dataset. \benchmark results demonstrate a clear advantage of the generalizable paradigm, particularly in relaxing the scene-specific limitation, enabling fast feed-forward inference on novel scenes, and achieving superior segmentation performance. We further introduce \dataall, a carefully curated 3DGS dataset comprising around 49K diverse indoor and outdoor scenes obtained from multiple sources, with which we demonstrate the generalizable approach could harness strong data priors. Our codes, benchmark, and datasets are available at \href{https://scenesplatpp.gaussianworld.ai/}{https://scenesplatpp.gaussianworld.ai/}.

%\yue{reformatted the abstract a bit to be more straightforward.}

% Recent advances in 3D computer vision have led to a strong focus on modeling geometry and visual appearance, while also striving to enable semantic understanding of complex scenes. Among various representations, Gaussian Splatting (3DGS) has emerged as a compelling choice due to its ability to simultaneously encode geometry, appearance, and semantic cues in a compact and dense format optimized from posed 2D images. In parallel, vision-language reasoning has shown great promise in bridging visual and geometric content with human language, enabling open-vocabulary and generalizable scene understanding. In this work, we bring these two directions together by introducing the first benchmark for vision-language scene understanding in the 3DGS domain. We introduce \dataall, a high-quality dataset of around \textbf{49K} curated indoor and outdoor 3D scenes represented as Gaussians. 
% Furthermore, we develop and evaluate a generalizable vision-language model that operates directly on 3DGS representations, setting a foundation for future research at the intersection of 3D representations and multimodal reasoning.
\end{abstract}
% the 2 kinds, with gradient and non-gradients, 

%-----------------------%
\section{Introduction}
\label{sec:introduction}
%-----------------------%

% \item Paragraph: Intro
% \begin{itemize}
%   \item 3DGS is the best 3D format.
%   \item Language features are very desirable.
%   \item Language features are orthogonal to 3DGS.
% \end{itemize}
The field of 3D computer vision is highly focused on capturing the geometry~\cite{WITKIN198117,Kazhdan2006,Hirschmuller2008StereoDepth,Newcombe2011KinectFusion,schoenberger2016sfm,Park2019DeepSDF,shi4568335city,ren2024bringing} and 
the visual appearance~\cite{10.5555/539405,791235,5226635,mildenhall2020nerf,kerbl3Dgaussians} of a scene, as well as understanding its content~\cite{8099499,wu2024ptv3,peng2023openscene,zhu2023pointclip,sarkar2025crossover}. Recently, 3D Gaussian Splatting (3DGS)~\cite{kerbl3Dgaussians} has emerged as the most desirable 3D representation due to its unique ability to jointly encode geometry, appearance, and scene understanding properties~\cite{Ye2024GSgroupping,Shi2024LEG,jiang2024open} in a compact form 
% Cut?
that can be effectively optimized from 2D posed images~\cite{ye2025gsplat}. 
Furthermore, vision-language reasoning represents the most promising direction for 3D scene understanding~\cite{rozenberszki2022language, ding2022pla,yang2024regionplc,ma2025cityloc,jiang2024open,wu2023mars,halacheva2025gaussianvlm}, as it bridges the visual and geometric attributes of a scene with the language we use to define, describe, and reason about concepts. Therefore, this work focuses exclusively on vision-language scene understanding with 3DGS.

% \item Paragraph: Types of vision-langauge 3DGS
% \begin{itemize}
%   \item Method G1: Gradient guided per-scene optimization
%   \item Method G2: Gradient-free per-scene optimization
%   \item Method G3: Generalizable single forward pass
% \end{itemize}
The most relevant methods for language Gaussian Splatting (LGS) can be categorized into three groups. %All of these groups assume access to a 3DGS representation of the scene as input, typically obtained via standard Gaussian Splatting optimization methods~\cite{kerbl3Dgaussians,Yu2024MipSplatting,kotovenko2025EDGS}. 
% Cut ?
% Furthermore, recent advances also allow reconstruction through generalizable single forward pass methods~\cite{charatan23pixelsplat,chen2024mvsplat,gslrm2024,ziwen2024llrm} or direct generation~\cite{roessle2024l3dg}.
%
The first two groups begin by extracting 2D features from all training images using vision-language foundation models (e.g., CLIP~\cite{Radford2021CLIP}). The first group then performs gradient-based per-scene optimization~\cite{Qin2024LangSplat,Zuo2024FMGS,zhou2024feature3DGS,Wu2024OpenGaussians,QU2024GOI}, assigning feature vectors to each 3D Gaussian primitive and optimizing them such that their renderings align with the corresponding 2D feature maps. The second group also operates on a per-scene basis but adopts an optimization-free approach~\cite{joseph2024GWfeatureBP,marrie2024ludvig,cheng2024occam}. Instead of iterative refinement, these methods directly lift 2D features to 3D primitives via weighted feature aggregation schemes. Lastly, we consider the generalizable approach. To date, only a single method has been trained on 2K indoor scenes~\cite{li2025SceneSplat7k}, learning to directly predict a vision-language feature vector for each 3D Gaussian primitive. Although 2D vision-language features are required during training, inference relies solely on 3D Gaussians, eliminating the need to prepare additional 2D feature maps. Furthermore, all features are produced in a single forward pass.

% \item Paragraph: Benchmarking issues \begin{itemize}
%   \item Issue 1: Mostly benchmarked on few scenes.
%   \item Why? Because it takes a lot of time and effort.
%   \item Consequence: Doesnt lead to general conclusions.
%   \item Issue 2: Mostly benchmarked near training camera views.
%   \item Why? 
%   \item Consequence: Doesnt necessarily work outside training views.
%   \item 3. Mostly benchmarked on 2D, and not 3D.
%   \item Why? 
%   \item Consequence: For true 3D scene understanding, we care how it performs in 3D. Which we cant know from 2D eval.
% \end{itemize}
While these methods represent important advances in integrating vision-language reasoning with 3DGS, their evaluation protocols still suffer from several critical limitations. First, most methods are evaluated on only a small number of selected scenes~\cite{Qin2024LangSplat,Zuo2024FMGS,zhou2024feature3DGS,Wu2024OpenGaussians,QU2024GOI,joseph2024GWfeatureBP,cheng2024occam}. As a result, such evaluations carry a high risk of producing scene-specific rather than generalizable conclusions. 
% % Cut?
% This is largely due to the substantial manual effort and computational cost required for a comprehensive evaluation.
% %to evaluate across a broader range of scenes. 
% %
Moreover, the absence of a standardized benchmark further limits systematic and comparable evaluation.
Second, most methods are evaluated close to the training views~\cite{Qin2024LangSplat,marrie2024ludvig,Wu2024OpenGaussians}. Consequently, there is no guarantee that the results generalize well to novel viewpoints. Finally, most evaluations are conducted in 2D rather than directly in 3D space~\cite{Qin2024LangSplat,Zuo2024FMGS,zhou2024feature3DGS,Wu2024OpenGaussians,QU2024GOI,joseph2024GWfeatureBP,marrie2024ludvig,cheng2024occam}. However, \textit{3D scene understanding fundamentally concerns performance in 3D space, which cannot be fully inferred from 2D projections}. 
Therefore, we introduce \benchmark, an evaluation benchmark for 3DGS vision-language scene understanding that includes 1060 scenes spanning 325 unique semantic classes. It evaluates performance in 3D for each segment of the scene, and we use it to assess representative methods from all three groups, as presented in~\cref{tab:benchmark_overview}.
% that includes a substantial number of scenes: 312 indoor scenes of ScanNet~\cite{dai2017scannet}, 50 indoor scenes of ScanNet++~\cite{yeshwanth2023scannet++}, 370 indoor scenes of Matterport3D~\cite{chang2017matterport3d}, and 328 outdoor scenes of HoliCity~\cite{zhou2020holicity}. In total, \benchmark comprises 1,060 scenes spanning 325 unique semantic classes. 
%
%\benchmark evaluates performance in 3D for each segment of the scene. Furthermore, we use the proposed benchmark to evaluate representative methods from each of the three aforementioned groups.

% \item Paragraph: Generalizable is better
% \begin{itemize}
%   \item No training time (per-scene optimization), given a foundation model.
%   \item The inference time is quick, only single forward pass.
%   \item Standard multi-view problems get solved with big data (occlusions; textureless areas; ...)
%   \item SceneSplat has better results than Occams which was used to label training data. Answer why this is (curve fitting analogy)
% \end{itemize}
Among the evaluated approaches, the generalizable 3D scene understanding paradigm demonstrates the strongest potential. It eliminates the need for substantial per-scene computation at inference and enables a single forward pass per 3D scene. In addition, this approach enables the application of 3D foundation models for scene understanding without the need for task- or scene-specific retraining. This concept aligns with established practices in 2D computer vision, where pretrained foundation models have been widely adopted~\cite{He_2016_CVPR,dosovitskiy2020vit,oquab2023dinov2,ravi2024sam2,Radford2021CLIP,liu2023llava}. Foundation models tailored to 3DGS have already emerged across key areas, including reconstruction~\cite{charatan23pixelsplat,chen2024mvsplat,gslrm2024,ziwen2024llrm}, scene understanding~\cite{li2025SceneSplat7k}, and generation~\cite{xiang2024structured,roessle2024l3dg}.
% % Cut ?
% Their ability to leverage large-scale data priors makes them particularly effective in mitigating common 3D challenges such as occlusions and textureless surfaces~\cite{Wang_2024_CVPR,wang2024vggsfm,wang2025vggt}.
% %
Lastly, benchmark results demonstrate the state-of-the-art performance of the generalizable approach, highlighting its capacity to leverage data priors and extract meaningful vision-language features—effectively mitigating the noise inherent in weak supervision.
%In addition, optimization-free per-scene methods consistently outperform optimization-based ones, which, together with the strong performance of generalizable methods, suggests that \textit{per-scene optimization is not necessary for effective vision-language reasoning in 3DGS}.

% \item Paragraph: SceneSplat dataset extension.
% \begin{itemize}
%   \item Summarize new scenes and new datast statistics.
%   \item Explain the new trained SceneSplat models (7k and outdoor).
%   \item Summarize benchmarking conclusions
% \end{itemize}
To facilitate the development of generalizable 3DGS scene understanding, we introduce \dataall, a carefully curated 3DGS dataset comprising diverse indoor and outdoor scenes collected from multiple sources. Existing large-scale 3DGS datasets primarily focus on single objects~\cite{ma2024shapesplat,liu24uco3d,wei20253daffordsplatefficientaffordancereasoning}, whereas scene-level datasets~\cite{li2025SceneSplat7k} remain limited in scale. To the best of our knowledge, \dataall represents the most extensive open-source dataset of complex, high-quality 3DGS reconstructions at the full-scene level. The preparation of the dataset required approximately 891 GPU days and considerable human involvement. Furthermore, we demonstrate that training a generalizable 3DGS scene understanding method on a larger subset from \dataall leads to improved performance, achieving state-of-the-art results on the benchmark.

% totaling 26.12B 3D Gaussian primitives. On average, the reconstructions achieve a PSNR of 27.56 dB and a depth error of 0.067 m. 

% \item Paragraph: Contributions
% \begin{itemize}
%   \item Meaningful benchmark
%   \item New dataset
%   \item Larger scale training (indoor and outdoor)
% \end{itemize}
Our contributions can be summarized as follows: 
\begin{itemize}[itemsep=0pt, parsep=0pt, topsep=0pt, leftmargin=*]
    \item We introduce \benchmark, the first \emph{benchmark} for systematically evaluating vision-language scene understanding methods in the 3DGS domain.
    \item We release \dataall, a high-quality, carefully curated 3DGS \emph{dataset} comprising 49K indoor and outdoor scenes.
    \item We scale up the training and evaluation of a generalizable VLM for 3DGS-based scene understanding across both indoor and outdoor environments, offering new insights.
\end{itemize}
% $\bullet$ 

\begin{table}[!t]
\resizebox{\textwidth}{!}{
\setlength{\tabcolsep}{4pt}% column sep.
\setlength{\extrarowheight}{0.1pt} % row height
\begin{tabular}{@{}lccr|lrrccc@{}}
\toprule
\textbf{Methods}          & 
\makecell{\textbf{Generalizable}}& \makecell{\textbf{Optimization-}\\\textbf{free}}&
\makecell[r]{\textbf{Runtime}\\\textbf{/ Scene}}&
\makecell[l]{\textbf{Benchmark}\\\textbf{Dataset}}& 
\makecell[r]{\textbf{Scene}\\\textbf{Number}}& \makecell[r]{\textbf{Class}\\\textbf{Number}}& \\
%& \makecell{\textbf{Feature}\\\textbf{Compression}}&
%\makecell{\textbf{2D/3D}\\\textbf{Evaluation}}& 
% &\makecell{\textbf{Train-Test}\\\textbf{Split/Novel wiew}}
\midrule
\rowcolor{tblgray}
\multicolumn{8}{l}{\textit{Per-Scene Optimization Methods}} \\
     LangSplat~\cite{Qin2024LangSplat}  & \crossmark & \crossmark  & 621 min & LERF~\cite{kerr2023lerf}, 3DOVS~\cite{liu2024weaklysupervised3dopenvocabulary} & 9 & 88  \\
 
    FMGS~\cite{Zuo2024FMGS}& \crossmark & \crossmark & 76 min & LERF~\cite{kerr2023lerf}, 3DOVS~\cite{liu2024weaklysupervised3dopenvocabulary} &11 &123 \\

    Feature3DGS~\cite{zhou2024feature3DGS}         & \crossmark & \crossmark & 235 min &Replica~\cite{straub2019replica}& 4 & 7  \\

    OpenGaussian~\cite{Wu2024OpenGaussians}           & \crossmark & \crossmark &297 min &LERF~\cite{kerr2023lerf}, \scannet~\cite{dai2017scannet}(10scenes) &14 & 76 \\

    GOI~\cite{QU2024GOI}& \crossmark & \crossmark & 252min  &Mip-NeRF360~\cite{barron2021mipnerf}, Replica~\cite{straub2019replica} & 8& 30 \\

\rowcolor{tblgray}
\multicolumn{8}{l}{\textit{Per-Scene Optimization-Free Methods}} \\
    
     \makecell[l]{Gradient-Weighted\\3DGS}~\cite{joseph2024GWfeatureBP} & \crossmark & \checkmark & 4 min & LERF~\cite{kerr2023lerf}, 3DOVS~\cite{liu2024weaklysupervised3dopenvocabulary} &9 &88 \\

    LUDVIG~\cite{marrie2024ludvig}  & \crossmark & \checkmark & 5.6 min & LERF~\cite{kerr2023lerf}, SPIn-NeRF~\cite{spinnerf},  NVOS~\cite{ren2022neural}  &21 &80 \\

    OccamLGS~\cite{cheng2024occam}  & \crossmark & \checkmark & 3.2 min& LERF~\cite{kerr2023lerf}, 3DOVS~\cite{liu2024weaklysupervised3dopenvocabulary} & 9 & 88\\

\rowcolor{tblgray}
\multicolumn{8}{l}{\textit{Generalizable Method}} \\
    
    SceneSplat~\cite{li2025SceneSplat7k} & \checkmark & \checkmark  & 0.24 min & \scannet\cite{dai2017scannet}, \scannetpp~\cite{yeshwanth2023scannet++},\matt~\cite{chang2017matterport3d} & 732 & 321 \\

\midrule
    \makecell[l]{\textbf{\benchmark}\\} & -- & -- & -- &  \makecell[l]{\textbf{\scannet}~\cite{dai2017scannet}, \textbf{\scannetpp}~\cite{yeshwanth2023scannet++},\\\textbf{\matt}~\cite{chang2017matterport3d}, \textbf{\holi}~\cite{zhou2021holicity}, }& \textbf{1060} & \textbf{325} \\

\bottomrule
\end{tabular}
}
\vspace{1mm}
\caption{\textbf{{Language Gaussian Splatting (LGS) Benchmark Overview}.} \textbf{Left:} Grouped methods and their properties. \textbf{Right}: Benchmark characteristics. Our proposed \benchmark benchmark evaluates the LGS methods at scale, across three indoor datasets and one outdoor dataset.}

\vspace{-8mm}
\label{tab:benchmark_overview}
\end{table}

%-----------------------%
\section{Related Works}
\label{sec:related_works}
%-----------------------%
\textbf{Gaussian Splatting Datasets.} Most existing 3DGS datasets remain object-centric, synthetic, or limited in scale and scope.
%—often lacking standardized benchmarks or access to full data. 
Moreover, recent works~\cite{charatan23pixelsplat,roessle2024l3dg,ziwen2024llrm,chen2024splatformer} showcase 3DGS generalization, yet do not fully release their datasets, thus limiting reproducibility and scalability.
% Moreover, recent works such as SplatFormer~\cite{chen2024splatformer}, L3DG~\cite{roessle2024l3dg}, LongLRM~\cite{ziwen2024llrm}, and PixelSplat~\cite{charatan23pixelsplat} emphasize generalization or multi-modality, yet do not fully release their datasets, thereby limiting reproducibility and scalability.
% 
Among publicly available datasets, ShapeSplat~\cite{ma2024shapesplat} transforms 65K object-level CAD models into 3DGS. 3DAffordSplat~\cite{wei20253daffordsplatefficientaffordancereasoning} focuses on functionality, providing 23K object instances labeled with affordances across 21 categories. 
%More recently, 
uCO3D~\cite{liu24uco3d} expands the object-centric landscape with over 1,000 annotated categories and 360$^\circ$ coverage.
%, showing benefits for both generative and discriminative 3D tasks.
%
Furthermore, scene-level 3DGS remains underexplored, with SceneSplat~\cite{li2025SceneSplat7k} being the only available dataset providing 6.8K indoor scenes to support open-vocabulary 3D understanding, though its scope is limited to the indoor domain.
In summary, existing datasets fall short of enabling generalizable, multi-modal learning across diverse scenes.
%, as most are restricted to object-level supervision or lack the scale and diversity necessary to train unified models for language-conditioned or zero-shot inference.
%
To address this, we introduce \ourdata, a large-scale dataset of 49K 3DGS scenes spanning indoor and outdoor environments, enriched with vision-language embeddings to support scene-level understanding.
% To facilitate generalizable 3DGS tasks such as scene understanding, we introduce \ourdata, a large-scale dataset comprising 49K 3DGS scenes spanning both indoor and outdoor environments, with vision-language embeddings.
%to facilitate scene understanding method. 

% This dataset enables the training of next-generation generalizable scene understanding methods in 3DGS and supports more comprehensive benchmarking of their performance.

%\ourdata is designed to support three learning paradigms—optimization-based, optimization-free, and generalizable modeling—and includes diverse scene-level annotations to facilitate robust multimodal learning. 
% By combining object and scene diversity, geometric and semantic signals, and large-scale real-world coverage, \ourdata establishes a comprehensive foundation for standardized evaluation and scalable training in the 3D Gaussian Splatting domain.

% Gaussian Splatting has become a widely adopted representation for 3D content, offering a compact and differentiable structure well-suited for learning-based tasks. This has led to a growing interest in large-scale 3DGS datasets.
% %constructing datasets beyond classic photorealistic rendering.  
% However, 

\textbf{Open-Vocabulary Scene Understanding.} 
% A lot of interests in open vocabulary in both 2D and 3D, 
% for gaussians splats, 
Recent advancements in open-vocabulary models have significantly expanded the capabilities of vision-language understanding. Foundation models~\cite{zhang2022dino, oquab2023dinov2,kirillov2023segment, ravi2024sam,radford2021learning, zhai2023sigmoid, tschannen2025siglip} enabled visual feature extraction at scale, effectively supporting tasks like detection, segmentation and open-vocabulary understanding.
Recent works have also introduced 2D foundation models into 3D by utilizing NeRF~\cite{mildenhall2021nerf} or 3DGS~\cite{kerbl20233d}. Existing 3DGS approaches fall into three categories: per-scene optimization-based~\cite{zhou2024feature3DGS, zheng2024gaussiangrasper, Qin2024LangSplat,guo2024semantic,QU2024GOI, Zuo2024FMGS, peng2023openscene, Wu2024OpenGaussians}, per-scene optimization-free~\cite{joseph2024GWfeatureBP, marrie2024ludvig,cheng2024occam} and generalizable~\cite{li2025SceneSplat7k}. Per-scene optimization-based LGS emerged first, with LERF~\cite{kerr2023lerf, rashid2023language} and LangSplat~\cite{Qin2024LangSplat} integrating 2D foundation model features into NeRF- and GS-based models, a design adopted by many later methods. Per-scene optimization-free LGS, such as OccamLGS~\cite{cheng2024occam}, LUDVIG~\cite{marrie2024ludvig} and Gradient Weighted 3DGS~\cite{joseph2024GWfeatureBP} further optimized this process by employing feature lifting, directly projecting all 2D features into 3DGS, reducing computation time from hours to seconds. 
SceneSplat~\cite{li2025SceneSplat7k} is the first generalizable LGS method, trained and evaluated directly on 3D data, achieving significant gains in open-vocabulary scene understanding with substantially faster inference on novel scenes.
However, most of these methods have been evaluated in constrained 2D settings, with a limited number of scenes and semantic classes.
We therefore introduce a comprehensive benchmark and use it to systematically evaluate recent language Gaussian Splatting methods.
%We thus comprehensively benchmark recent LGS methods on larger scale under the unified settings.
%Yet, these methods are all evaluated in very limited scale of scenes like LERF~\cite{kerr2023lerf} and 3D OVS~\cite{liu2024weaklysupervised3dopenvocabulary} or different settings like chosen scenes and limited classes under ScanNet~\cite{dai2017scannet, scannet200}, ScanNet++~\cite{yeshwanth2023scannet++}. 

\textbf{3D Generalizability.} Large-scale training and generalization in 2D have become increasingly mature~\cite{Radford2021CLIP, tschannen2025siglip, lyu2025realrag, lyu2024unibind, ravi2024sam, ravi2024sam2, zhong2025omnisam}. In contrast, achieving semantic and language generalization in 3D remains challenging and relatively underexplored~\cite{chen2024splatformer, PointMAE, zhao2021point, wu2022point, wu2024ptv3, lee2025mosaic3d, wu2025sonata}. One major obstacle is the limited quantity and fragmented nature of 3D datasets, as no single dataset provides the needed combination of (i) large-scale coverage, (ii) multi-scale diversity (from object-centric to indoor and outdoor scenes), and (iii) rich multimodal information (geometry, appearance, and semantics). As a result, even state-of-the-art 3D models struggle to generalize broadly. Mosaic3D-5.6M compiles over 30K RGB-D scenes with 5.6 million mask-text pairs by leveraging 2D vision-language models, yielding an significant resource for open-vocabulary 3D segmentation~\cite{lee2025mosaic3d}. The SceneSplat-7K dataset supports open-vocabulary training but with limited scale~\cite{li2025SceneSplat7k}. These efforts represent early steps toward 3D foundation models, aiming to enable robust, optimization-free, and generalizable 3D reasoning across vision-language tasks. 
Therefore, we introduce \dataall, a large and comprehensive dataset designed to unlock scaling laws in generalizable 3DGS tasks. 

%-----------------------%
\section{Dataset}

\label{sec:dataset}

\begin{table}[!t]
    \centering
    % \footnotesize
    \small
    \resizebox{\linewidth}{!}{
    %\scalebox{0.9}{
    \setlength{\tabcolsep}{2pt}% column sep.
    \setlength{\extrarowheight}{0.1pt} % row height
    \begin{tabular}{lcccccr}
        \toprule
        \textbf{Property} & \textbf{Scene\-Splat‑7K~\cite{li2025SceneSplat7k}} & \textbf{DL3DV‑10K~\cite{ling2024dl3dv}} &
        \textbf{HoliCity~\cite{zhou2021holicity}} & \textbf{Aria‑ASE~\cite{avetisyan2024scenescript}} & \textbf{Crowdsourced} &        \textbf{Total} \\
        \midrule
        % ---------- DIVIDER ROW ----------
        \rowcolor{tblgray}
        \multicolumn{7}{l}{\textit{Data Statistics}} \\
        Raw Scenes        & 9114   & 10K & 6\,253 & 25K & 603 & 48\,856 $\sim$49K \\
        GS Scenes         & 7\,916     & 6\,376    & 5\,940  & 25K & 603   & 45\,835 $\sim$46K   \\
        Data Type         & Indoor  & Indoor/Outdoor & Outdoor   & Indoor  & Indoor/Outdoor &  Indoor/Outdoor        \\
        RGB Frames        & 4.72M     &  3.09M   & 48K      & 14.79M    & N/A  & 22.65M       \\
        Avg. GS / Scene        & 1.42M     &  1.48M   & 800K      & 1.27M    & 1.02M  & 1.26M          \\ % 1274K->1.27M
        3DGS GPU Time [h]   & 3.59K     &  2.1K   &  0.79K    & 6.25K   & N/A  & 12.73K       \\
        Scenes w/ VLM Embedding   & 6\,121     &  -   &  5\,940    & -   & -  &12\,061    \\%12.1K
        Embedding GPU Time [h]   & 8\,161     &   -  &  495    & -   & -  & 8\,656       \\
        Storage [TB]      & 2.76    & 2.91   & 0.26      & 2.2    &  0.23   & 8.36          \\
        Total 3DGS   & 11.27B   &  9.42B   & 4.75B      & 3.18B    & 615M   & 29.24B          \\%Holicity Total3DGS 4752M ->4.75B
        % ---------- DIVIDER ROW ----------
        \rowcolor{tblgray}
        \multicolumn{7}{l}{\textit{Evaluation Metrics (Avg.)}} \\
        % ---------- EVALUATION METRICS ----------
        PSNR [dB] $\uparrow$        & 29.64 & 25.70 & 29.85 & 27.32 & N/A & 27.83 \\
        Depth Loss [m] $\downarrow$& 0.035 & - & 0.010 & 0.082 & N/A & 0.061 \\
        SSIM $\uparrow$        & 0.897 & 0.845 & 0.921 & 0.906 & N/A & 0.898 \\
        LPIPS $\downarrow$     & 0.212 & 0.203 & 0.122 & 0.230 & N/A & 0.209 \\
        \bottomrule
    \end{tabular}
    }
    \vspace{2mm}

    \caption{\textbf{\ourdata Statistics and Quality Metrics.}
    Our proposed dataset includes outdoor and indoor scenes drawn from established datasets and publicly available sources, including our own collected data. In total, it comprises approximately \qty{49}{\K} raw scenes, \qty{46}{\K} curated 3D Gaussian Splatting (3DGS) scenes, and \qty{29.24}{\B} Gaussian splats. Generating all 3DGS required \num{\sim530} GPU days on an NVIDIA L4 GPU. Appearance and geometry reconstruction quality are evaluated using four metrics: PSNR, SSIM, LPIPS and depth $\ell_1$ error. Additionally, \qty{12}{\K} of the 3DGS scenes are annotated with per-primitive vision–language embeddings to support the training of generalizable scene understanding methods. Computing these embeddings further required \num{\sim361} GPU days.
    }
    % peak signal-to-noise ratio (PSNR)
    % structural similarity index (SSIM)
    % learned perceptual image patch similarity (LPIPS)

    \label{tab:dataset_statistic}
    %\vspace{-2mm}
    \vspace{-4mm}
    
\end{table}

% \begin{table*}[htbp]
%     \centering
%         \setlength{\tabcolsep}{2pt}
%         \begin{tabular}{l|ccccc|c}
%             \toprule[0.95pt]
%         \textbf{Property} & \textbf{SceneSplat-7K}&\textbf{DL3DV-10K} & \textbf{HoliCity} & \textbf{Aria-ASE} & \textbf{User-shared}&  \textbf{Total} \\
%         \midrule[0.6pt]
%         \textbf{Raw Scenes} & 7K & 10k & 6253 & 25K & 600 &  $\sim$49k \\
%         \textbf{GS Scenes} & - & - & 5940 & 25K &   \\
%         \textbf{Data Type} & - & - & Outdoor & Indoor &  \\
%         \textbf{RGB Frames} & - & - & - & - \\
%         \textbf{GS per scene} & - & - & - & - \\
%         \textbf{Total GS} & - & - & - & - \\
%         \textbf{GPU Time} & - & - & - & - \\
%         \textbf{Storage} & - & - & - & - \\
        
%         \midrule[0.6pt]
        
%         \textbf{PSNR} & - & - & - & -\\
%         \textbf{Depth Loss} & - & - & - &-\\
%         \textbf{SSIM} & - & - & - & - \\
%         \textbf{LPIPS} & - & - & - & -\\

%         \bottomrule[0.95pt]
%         \end{tabular}
%         \vspace{-2mm}
%     \caption{
%             \textbf{Dataset Statistics.} The proposed dataset includes outdoor datasets...(\mengjiao{do we include the human-collected data here?})
%             The dataset contains ... scenes and ... 3DGS. Constructing this dataset requires computational resources equivalent to ... GPU-days on one NVIDIA L4 GPU. ..achieves high-fidelity reconstruction quality with an average PSNR ... dB.}
%         \label{tab:dataset_statistic}
%     \vspace{-3mm}
% \end{table*}

We present \ourdata, a large-scale 3D Gaussian Splatting dataset comprising approximately \qty{49}{\K} raw scenes and \qty{46}{\K} curated 3DGS scenes, aggregated from multiple established sources, including SceneSplat-7K~\cite{li2025SceneSplat7k}, DL3DV-10K~\cite{ling2024dl3dv}, HoliCity~\cite{zhou2021holicity}, Aria Synthetic Environments~\cite{avetisyan2024scenescript}, and our own crowdsourced data. This diverse dataset consists of both indoor and outdoor environments, spanning from rooms, apartments to streets. To support 3DGS scene understanding model training, \qty{12}{\K} scenes are further enriched with per-primitive vision-language embeddings extracted with state-of-the-art vision-language models~\cite{tschannen2025siglip}. The comprehensive statistics of the introduced dataset are presented in~\cref{tab:dataset_statistic}.

% The dataset in total contains \qty{26.12}{\B} Gaussian splats optimized from \qty{22.4}{\M} RGB frames, with an average of \qty{1.25}{M} Gaussians per scene.

% The creation of \ourdata required substantial resources—approximately \num{500} GPU-days for 3DGS optimization and an additional \num{361} GPU-days for embedding generation—resulting in a comprehensive dataset totaling \qty{7.87}{\tera\byte} of storage.

\begin{figure}[!t]
    \centering
    \vspace{-2mm}
    \includegraphics[width=0.9\columnwidth]{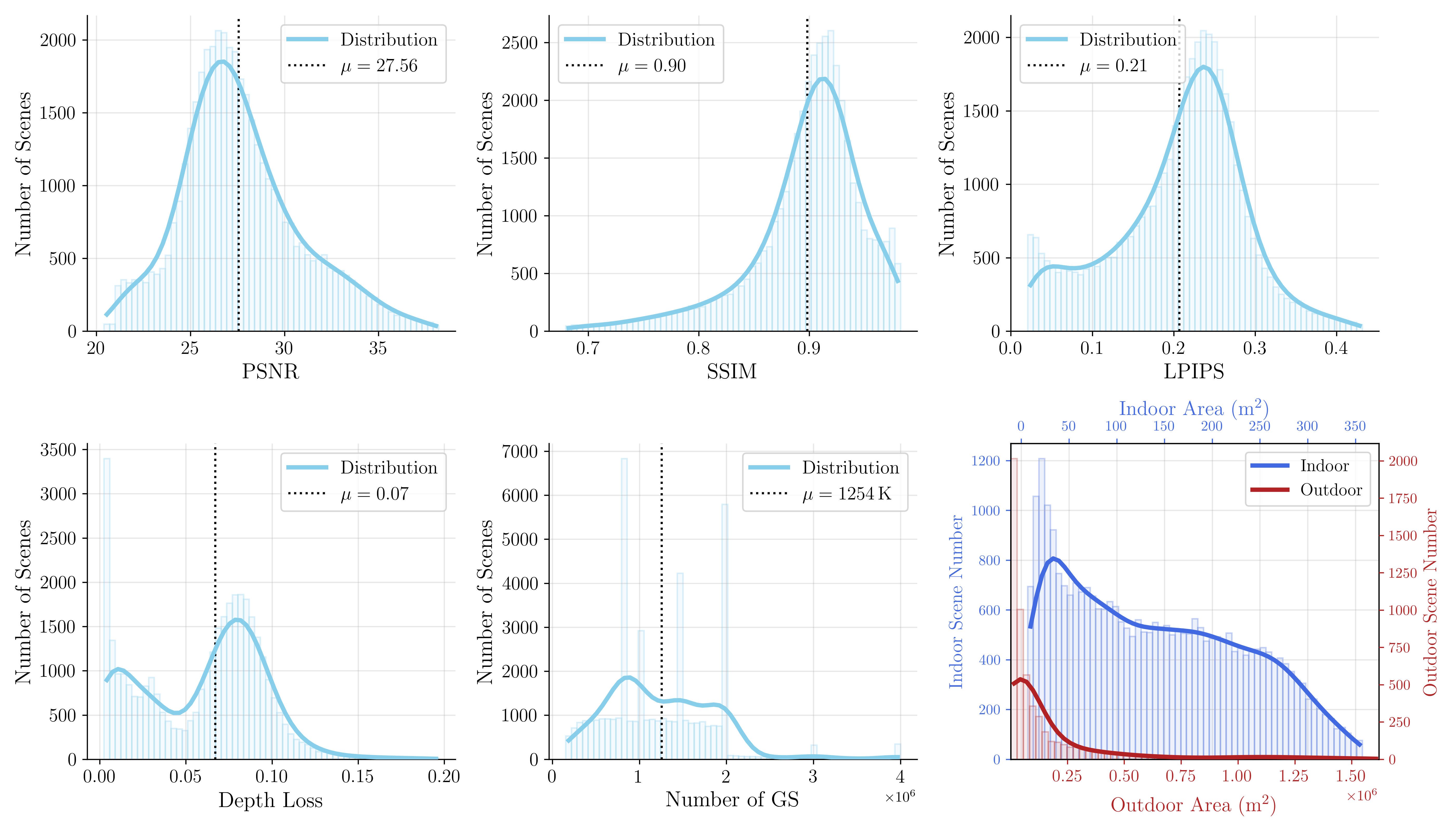}
    \caption{\textbf{Appearance, Geometry, and Scale Statistics of the \dataall Dataset.}
    Distributions of photometric (PSNR, SSIM, LPIPS) and geometric (depth $\ell_{1}$) reconstruction errors show consistently high‐quality renders across scenes, while the wide spread in total Gaussian number and indoor/outdoor scene floor area demonstrates the dataset’s diversity. The curves are convolved from the bucket values and vertical dotted lines mark the mean of each metric.}
    \label{fig:dataset_stats}
    \vspace{-3mm}
    % \mengjiao{Nikola suggests to change subgraph5- Number of GS to histogram or bar graph, as the number of GS is discrete.}
    % \nikola{I suggested to change every subgraph because statistics are usually prsented as histograms. Because of discrete bin counts. Even though I actually personally prefer this way, its the usual practice.}
    % \yue{let's discuss, I prefer the smoothed version.}
\end{figure}

\cref{fig:dataset_stats} visualizes the distribution of evaluation metrics of our \dataall dataset.  
Across \qty{\sim49}{\K} 3DGS scenes, the mean photometric quality is \qty{27.8}{\decibel} PSNR and \num{0.90} SSIM, with a perceptual LPIPS of \num{0.20}—values that are of high quality renders. 
Importantly, the geometry reconstruction is equally reliable, with a mean depth $\ell_1$ error of \qty{0.061 }{\metre}. The distribution of the per-scene Gaussian number spans two orders of magnitude, indicating the dataset’s complexity. The scene footprint areas also show clear diversity: indoor environments cluster all across \qtyrange{25}{250}{\square\metre}, whereas outdoor scenes extend over a square kilometer, with a long-tailed distribution. This diversity in spatial extent, together with high-quality appearance and geometry, underpins the suitability of the dataset for training a generalizable 3DGS scene understanding model.

\subsection{Data Collection and Processing}
\label{sec:data_collecting}

To ensure high-quality 3DGS scenes, we implement several quality control measures throughout the optimization pipeline. Starting with the training views, we select scenes with at least 400 frames to ensure sufficient multi-view coverage. When depth information is available, we initialize the Gaussians at fused point cloud locations and apply depth supervision to maximize geometry quality. We use gsplat~\cite{ye2025gsplat} for 3DGS optimization. We filter blurry frames using the variance of the Laplacian as a sharpness metric. To efficiently compress the 3DGS scene, we employ the Markov Chain Monte Carlo strategy~\cite{kheradmand20253d} and add opacity and scale regularization. Once optimized, the dataset supports filtering 3DGS scenes based on PSNR and depth quality before using them as inputs for model training. We introduce the processing of each data source in the following.

\boldparagraph{\dataindoor} contains 7,916 curated 3DGS indoor scenes derived from ScanNet~\cite{dai2017scannet}, ScanNet++~\cite{yeshwanth2023scannet++}, Replica~\cite{straub2019replica}, Hypersim~\cite{roberts2021hypersim}, 3RScan~\cite{wald2019rio}, ARKitScenes~\cite{baruch2021arkitscenes}, and Matterport3D~\cite{chang2017matterport3d}.We further enrich over 6,000 of these scenes with vision–language embeddings to support open-vocabulary scene queries.

\boldparagraph{Aria Synthetic Environments} (ASE) is a large-scale synthetic dataset of procedurally-generated multi-room indoor scenes, each populated with 3D object models drawn from a digital asset library. We select the first \qty{25}{K} scene subset of it. For our 3DGS optimization, we undistort the fisheye image and depth frames and apply the devignetting mask for brightness correction. The sensor depths are fused and transformed to yield point clouds for the 3DGS initialization.

\boldparagraph{DL3DV-10K} dataset covers 65 everyday environments, featuring both indoor and outdoor locations. It includes 10,510 high-quality videos captured by mobile devices and drones, along with sparse COLMAP reconstructions. For 3DGS training, we use an evenly sampled set of every 10 views for novel view evaluation, while the remaining views are used for training.

% We utilize the proposed COLMAP camera pose and sparse depth information to initialize Gaussian splatting. Since the scene scale varies significantly, we use the default Gaussian to accommodate a wide range of Gaussian values. We filter out any data with a PSNR below 21, with an average PSNR of 25.7. Additionally, we selected a 2K subset to perform COLMAP dense reconstruction, achieving a PSNR of 29.14 and an average of 2.5 million Gaussians.

% dataset covers 65 everyday environments, including both indoor and outdoor locations. We utilize the proposed COLMAP camera pose and sparse depth information to initialize Gaussian splatting. Since the scene scale varies significantly, we use the default Gaussian to accommodate a wide range of Gaussian values. We filter out any data with a PSNR below 21, with an average PSNR of 25.7. Additionally, we selected a 2K subset to perform COLMAP dense reconstruction, achieving a PSNR of 29.14 and an average of 2.5 million Gaussians.

\boldparagraph{HoliCity} is a city-scale 3D dataset comprising 6,300 real-world panoramas, each accurately aligned to a downtown London CAD model covering over \qty{20}{\kilo\square\metre}, providing depth frames and semantic labels. We used eight perspective views from each panorama generated by evenly sampling the yaw angles at 45° intervals to optimize the 3DGS. 

\boldparagraph{Crowdsourced Data.}
To diversify our dataset, we collected high-quality 3DGS scenes from various sources, including Polycam and Sketchfab. We have also distributed our questionnaire to the community to collect data.

\subsection{Vision-Language Embedding Collection}
Vision-language embeddings are extracted from the frames used to optimize the 3DGS. Unlike existing pretraining methods that align 3D primitives with text embeddings from vision-language models, we align each Gaussian directly in the image‐embedding space, thereby preserving richer latent semantics. This approach avoids the information loss inherent in textual descriptions. 
% and overcomes the limitations of visual captioning models that struggle to capture all scene content aspects.

We follow the fusion strategy in~\cite{li2025SceneSplat7k} and employ a dynamic weighting mechanism that adaptively combines three distinct features: global context from the entire frame, local features with background, and masked features without background. This mechanism automatically balances contributions based on each segment's contextual relationship—emphasizing background-inclusive features for integrated objects (e.g., keyboard with monitor) and object-specific features for isolated objects. This adaptive approach provides more nuanced semantic understanding than fixed weighting strategies.

% Further details are given in~\cite{li2025SceneSplat7k}. \nikola{Maybe mention more clearly that our data processing is based on SceneSplat?} 

%-----------------------%
%-----------------------%
\section{Benchmark}
\label{sec:Benchamrk}
%-----------------------%

% \subsection{Problem Formulation}

We formulate the assigning of Gaussian language feature as follows. A scene is represented by a set of \(N\) Gaussian primitives \(\mathcal{G} = \{G_j\}_{j=1}^{N}\), where each \(G_j \in \mathbb{R}^{59}\) encodes its position, covariance, color, and opacity. For every primitive we compute a language embedding \(E_j^{G} \in \mathbb{R}^{d}\), obtained by per-scene processing or by inference with a learned encoder. The user provides a set of free-form text queries \(C = \{c_i\}_{i=1}^{n}\). A frozen text encoder \(P_T\) (e.g., CLIP) maps each query to an embedding \(E_i^{T} = P_T(c_i)\). Each Gaussian is labeled with the query whose embedding is most similar to its
own:
\begin{equation}
\text{label}(G_j) \;=\;
\arg\max_{i}\; \operatorname{cos}\bigl(E_j^{G},\, E_i^{T}\bigr).
\end{equation}
This label field assigns every primitive to the most relevant query.
% \martin{$sim(\cdot)$ is undefined.}

\subsection{Benchmark settings}
\label{sec:benchmark_settings}
\boldparagraph{Evaluation Protocols.} 
We evaluate 9 methods from 3 categories on our \benchmark benchmark. We report two key metrics in 3D: foreground mean Intersection over Union (f-mIoU) and foreground mean accuracy (f-mAcc), which address object size imbalances and reduce viewpoint dependency compared with traditional 2D-only evaluation. We unified the evaluation pipeline by first converting the language fields into a unified format $\mathcal{E}_{G} = \{E_j\}_{j=1}^N$, then computing the cosine similarity between each language feature and the queried text embeddings. Each Gaussian primitive is assigned the semantic label corresponding to the most similar text embedding. 
For every ground-truth 3D point used for evaluation, we identify its K-nearest Gaussian neighbors (K=25) and assign the majority-voted label as the semantic prediction for that  point. We exclude floor, wall, and ceiling classes as background for indoor scenes and the sky class for outdoor scenes. To make our evaluation more complete, we additionally evaluate semantics on 2D renders, following the protocol of LERF~\cite{kerr2023lerf}. The 2D evaluation is conducted on subsets of the benchmark datasets with two key 2D metrics: mean Intersection over Union (mIoU) and mean accuracy (mAcc) on all classes (background and foreground classes). All runtime measurements were obtained on NVIDIA RTX A6000 GPUs, except for FMGS, whose training requires 50–80 GB of GPU memory; its runtime was therefore measured on NVIDIA H200 GPUs.
% then computing the similarity with the queried text embedding and selecting the pair with the highest similarity. K-nearest neighbors are used to vote predictions from Gaussian splats to ground truth label positions. We exclude floor, wall, and ceiling classes as background for indoor scenes and the sky class for outdoor scenes. 

\boldparagraph{Benchmark Dataset.} 
We evaluate the methods' performance on indoor scenes using ScanNet (with 20 and 200 classes), ScanNet++ (100 classes), and Matterport3D (21 classes). Queries cover a wide range of object granularity—from small items like a mouse to larger objects such as a bed—reflecting the method’s capability to varying scales understanding. HoliCity (4 classes) is employed to evaluate the performance on outdoor scenes, emphasizing large structures under complex surrounding environment. Performance results are reported on randomly sampled 10-scenes subsets for each of the four benchmarks across all methods. For methods demonstrating strong efficiency, we conduct a full evaluation on the complete validation set.

\boldparagraph{Baseline Methods.} We categorize the methods based on their optimization paradigm and generalizability to novel scenes.
% shorter version
\boldparagraph{\textit{(i)} Per-Scene Optimization-based Methods} assign learnable attributes to Gaussian splats and optimize them via backpropagation. However, rendering additional features with alpha compositing is computationally expensive and requires per-scene optimization. 
\boldparagraph{\textit{(ii)} Per-Scene Optimization-Free methods} use a training-free approach, lifting 2D features into 3D space as a weighted sum of 2D representations, requiring only a single forward rendering pass. 
\boldparagraph{\textit{(iii)} Generalizable Method} SceneSplat trains a feed-forward 3DGS encoder, learning to predict representations aligned with image embeddings. We further extend its training based on our dataset. SceneSplat (Pseudo Label) denotes the VLM embeddings used for its vision-language pretraining. We refer to the supplement for details on running each baseline.

\begin{figure}[ht]
  \centering
\includegraphics[width=0.9\linewidth]{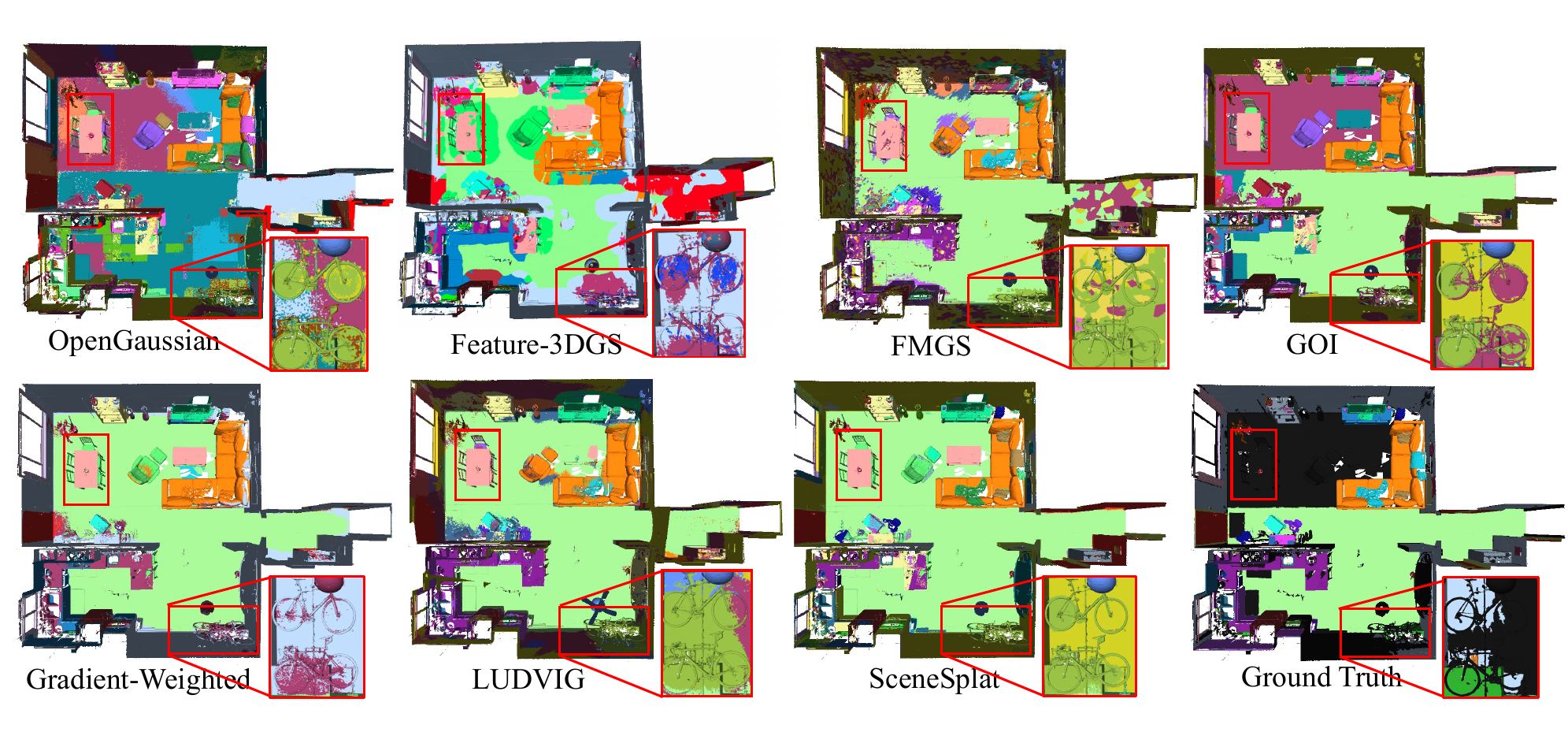}

  \caption{\textbf{Qualitative Results of Zero-Shot 3D Semantic Segmentation.} The semantic classes "bicycle" and "kitchen table" are highlighted, which are not labeled in Ground Truth.}
  \label{fig:vis_sem_seg}
\end{figure}

\begin{table}[!t]
  \centering
  \small
  %\scalebox{0.8}{
   \resizebox{\linewidth}{!}{%
    \setlength{\tabcolsep}{2pt}%
    \begin{tabular}{@{}
                  l
                  cc   % 10 scenes: f-mIoU | f-mAcc
                  cc   % all_val:   f-mIoU | f-mAcc
                  cc
                  cc
                  % c    % Pre-process
                  % c    % Training
                  @{}}
    \toprule
    \multirow{2.8}{*}{\textbf{Method}} &
      \multicolumn{2}{C{3cm}}{\makecell{\textbf{ScanNet20}\\\textbf{(10 scenes)}}} &
      \multicolumn{2}{C{3cm}}{\makecell{\textbf{ScanNet20}\\\textbf{(312 scenes)}}} &
      \multicolumn{2}{C{3cm}}{\makecell{\textbf{ScanNet200}\\\textbf{(10 scenes)}}} &
      \multicolumn{2}{C{3cm}}{\makecell{\textbf{ScanNet200}\\\textbf{(312 scenes)}}} \\
    \cmidrule{2-3}\cmidrule{4-5}\cmidrule{6-7}\cmidrule{8-9}
      & f–mIoU & f–mAcc & f–mIoU & f–mAcc & f–mIoU & f–mAcc & f–mIoU & f–mAcc \\
    \midrule
    
    \rowcolor{tblgray} 
    \multicolumn{9}{l}{\textit{Per-Scene Optimization Methods}} \\

    LangSplat~\cite{Qin2024LangSplat} & 0.0153     & 0.0741    & -    & -    & 0.0057  & 0.0106 & - &- \\
    OpenGaussian~\cite{Wu2024OpenGaussians}  & 0.1186   &  0.2340    & - &-    & 0.0545  &  0.1086 &- &-   \\
    Feature-3DGS~\cite{zhou2024feature3DGS}  & 0.1711    & 0.2512    & -    & -    & 0.0541  & 0.0876 &- &-  \\
    FMGS~\cite{Zuo2024FMGS} & 0.0916  & 0.2028   & 0.1169    & 0.2684    &  0.0641  & 0.1374 & 0.0579 & 0.1443 \\
    GOI~\cite{QU2024GOI}  & 0.2295   & 0.4638   & 0.2115    & 0.4286    & 0.1193   & 0.2297   & 0.1001 & 0.2240\\

    \rowcolor{tblgray}
    \multicolumn{9}{l}{\textit{Per-Scene Optimization-Free Methods}} \\
     Gradient-Weighted 3DGS~\cite{joseph2024GWfeatureBP}   & \textbf{0.4184}    & 0.5174   & 0.3573    & 0.4638    & 0.1638    & 0.2502   & 0.0601  & 0.1100 \\
    LUDVIG~\cite{marrie2024ludvig}  &0.1406    & 0.2885   & 0.1337    &  0.2969    & 0.0887   & 0.1833  & 0.0602 &  0.1701\\
    OccamLGS~\cite{cheng2024occam}  & 0.3559    & 0.4587   & 0.2308    & 0.4148   & 0.1936  & 0.2458 & 0.1204 & 0.2503   \\
    \textcolor{gray}{SceneSplat~\cite{li2025SceneSplat7k} (Pseudo Label)} & \textcolor{gray}{0.3521} & \textcolor{gray}{0.5210} & \textcolor{gray}{0.3500} & \textcolor{gray}{0.5550} & \textcolor{gray}{\textbf{0.2520}} & \textcolor{gray}{0.3038} & \textcolor{gray}{\textbf{0.2280}} & \textcolor{gray}{\textbf{0.3590}} \\
    
    \rowcolor{tblgray}
    \multicolumn{9}{l}{\textit{Generalizable Method}} \\
   SceneSplat~\cite{li2025SceneSplat7k}  & 0.2968   & \textbf{0.5711}    & \textbf{0.3540}  &  \textbf{0.5780} & 0.1387 & \textbf{0.4198} & 0.1648 & 0.3573 \\
    \bottomrule
  \end{tabular}
}
  \vspace{1mm}
 \caption{\textbf{Zero-shot 3D Semantic Segmentation Experiments on ScanNet20~\cite{dai2017scannet} (20 classes) and ScanNet200~\cite{dai2017scannet} (200 classes)}. All methods are evaluated on a 10-scene mini-validation set, with the 312-scene full validation set run only for selected methods due to runtime limitations (\cref{tab:benchmark_overview}).}
 \label{tab:3D_eval_scannet}
  % \vspace{-1cm}
 \vspace{-3mm}
\end{table}

%\Comparison of methods on ScanNet (10 scenes \& full validation set).

%TODO
%split the table 
%part1 only on 10 scenes
%part2 full_val_set -> show the story, the performance drop on full_val_set

\begin{table*}[!t]
  \centering
  \small
  \resizebox{\linewidth}{!}{%
  % \scalebox{0.88}{
    \setlength{\tabcolsep}{2pt}%
    \begin{tabular}{@{}
                  l
                  cc   % 10 scenes: f-mIoU | f-mAcc
                  cc|   % all_val:   f-mIoU | f-mAcc
                  cc
                  cc
                  % c    % Pre-process
                  % c    % Training
                  @{}}
    \toprule
    \multirow{2.8}{*}{\textbf{Method}} &
      \multicolumn{2}{C{3cm}}{\makecell{\textbf{ScanNet++}\\\textbf{(10 scenes)}}} &
      \multicolumn{2}{C{3cm}|}{\makecell{\textbf{ScanNet++}\\\textbf{(50 scenes)}}} &
      \multicolumn{2}{C{3cm}}{\makecell{\textbf{\matt}\\\textbf{(10 scenes)}}} &
      \multicolumn{2}{C{3cm}}{\makecell{\textbf{\matt}\\\textbf{(370 scenes)}}} \\
    \cmidrule{2-3}\cmidrule{4-5}\cmidrule{6-7}\cmidrule{8-9}
      & f–mIoU & f–mAcc & f–mIoU & f–mAcc & f–mIoU & f–mAcc & f–mIoU & f–mAcc \\
    \midrule
    
    \rowcolor{tblgray}
    \multicolumn{9}{l}{\textit{Per-Scene Optimization Methods}} \\

    LangSplat~\cite{Qin2024LangSplat} & 0.0131     & 0.0332    & -    & -    &  0.0219     & 0.1257  & - &- \\
    OpenGaussian~\cite{Wu2024OpenGaussians}  & 0.0695    & 0.1186   & - &-    & 0.1387      & 0.2698 &- &-   \\
    Feature-3DGS~\cite{zhou2024feature3DGS}  &  0.0798    & 0.1765     & -    & -    &  0.3043      & 0.4732   &- &-  \\
    FMGS~\cite{Zuo2024FMGS} & 0.0984  & 0.1892   & 0.1004    & 0.2300    & 0.2105     & 0.3664  & -  & - \\
    GOI~\cite{QU2024GOI}  & 0.1593    & 0.2790  & 0.1631    & 0.3296     &  0.1631      & 0.3296   & 0.1652      & 0.3130  \\

    \rowcolor{tblgray}
    \multicolumn{9}{l}{\textit{Per-Scene Optimization-Free Methods}} \\
     Gradient-Weighted 3DGS~\cite{joseph2024GWfeatureBP}  & 0.0925  & 0.1391 & 0.0900    & 0.1328      &  \textbf{0.3083}      & 0.4586      & 0.2762      & 0.4187    \\
    LUDVIG~\cite{marrie2024ludvig}  &  0.0995    & 0.2099   & 0.1099    &  0.2547    & 0.2348 & 0.3888 & 0.1960 & 0.3738\\
    OccamLGS~\cite{cheng2024occam}  & 0.1580    & 0.2861   & 0.1502    & 0.3312    & 0.1803 & 0.3263 & 0.1725 & 0.3151   \\
    \textcolor{gray}{SceneSplat~\cite{li2025SceneSplat7k} (Pseudo Label)} & \textcolor{gray}{\textbf{0.2340}} & \textcolor{gray}{0.3709} & \textcolor{gray}{0.2243} & \textcolor{gray}{0.4667}  &   \textcolor{gray}{0.2392} & \textcolor{gray}{0.4368} & \textcolor{gray}{0.2748} & \textcolor{gray}{0.4384}    \\
    
    \rowcolor{tblgray}
    \multicolumn{9}{l}{\textit{Generalizable Method}} \\
   SceneSplat~\cite{li2025SceneSplat7k}  & 0.2263  & \textbf{0.4916}   & \textbf{0.2836}   &  \textbf{0.4992}& 0.2732 & \textbf{0.5379} & \textbf{0.3384} & \textbf{0.5745} \\
    \bottomrule
  \end{tabular}
}
 \vspace{1mm}
 \caption{\textbf{Zero-Shot 3D Semantic Segmentation Experiments On \scannetpp~\cite{yeshwanth2023scannet++} and \matt~\cite{chang2017matterport3d}}. All methods are evaluated on a 10-scene mini-validation set, with the full validation set evaluated only for selected methods due to runtime limitations (\cref{tab:benchmark_overview}).}
 \label{tab:3D_eval_scannetpp_matterport}

\end{table*}

%\hl{Comparison of methods on ScanNet++ and Matterport3D (10 scenes \& full validation set).}

\begin{table}[htbp]
  \centering
    \setlength{\tabcolsep}{16pt}% column sep.

  \small
  \resizebox{\linewidth}{!}{
\begin{tabular}{@{}l
  cc
  cc |
  cc
  cc
  @{}}
    \toprule
    \multirow{2.4}{*}{\textbf{Method}} &
      \multicolumn{2}{c}{\textbf{ScanNet20 (10 scenes)}} &
      \multicolumn{2}{c|}{\textbf{ScanNet200 (10 scenes)}} &
      \multicolumn{2}{c}{\textbf{ScanNet++ (10 scenes)}} \\
    \cmidrule(lr){2-3}\cmidrule(lr){4-5}\cmidrule(lr){6-7}
    & mIoU & mAcc & mIoU & mAcc & mIoU & mAcc \\
    \midrule
    \rowcolor{tblgray}
    \multicolumn{7}{l}{\textit{Per‑Scene Optimization Methods}} \\
    LangSplat~\cite{Qin2024LangSplat} & 0.0193 & 0.0700 & 0.0133 & 0.0272 & 0.0111 & 0.0272 \\
    OpenGaussian~\cite{Wu2024OpenGaussians} & 0.1958 & 0.3294 & 0.0740 & 0.1273 & 0.1290 & 0.2133 \\
    Feature-3DGS~\cite{zhou2024feature3DGS} & \textbf{0.5054} & \textcolor{gray}{0.6474} & \textcolor{gray}{0.1867} & \textcolor{gray}{0.2798} & 0.1605 & 0.2368 \\
    FMGS~\cite{Zuo2024FMGS} & 0.1413 & 0.2629 & 0.0848 & 0.1825 & 0.1134 & 0.2244 \\
    GOI~\cite{QU2024GOI} & 0.2866 & 0.4536 & 0.1134 & 0.1806 & 0.1635 & 0.2767 \\
    \rowcolor{tblgray}
    \multicolumn{7}{l}{\textit{Per‑Scene Optimization‑Free Methods}} \\
    Gradient-Weighted 3DGS~\cite{joseph2024GWfeatureBP} & \textcolor{gray}{0.4533} & 0.5785 & 0.1639 & 0.2367 & 0.1210 & 0.1760 \\
    LUDVIG~\cite{marrie2024ludvig} & 0.1673 & 0.3058 & 0.0958 & 0.1828 & 0.1239 & 0.2588 \\
    OccamLGS~\cite{cheng2024occam} & 0.2937 & 0.4645 & 0.1300 & 0.2146 & \textcolor{gray}{0.1670} & \textcolor{gray}{0.2866} \\
    \rowcolor{tblgray}
    \multicolumn{7}{l}{\textit{Generalizable Method}} \\
    
    SceneSplat~\cite{li2025SceneSplat7k} & 0.4458 & \textbf{0.6727} & \textbf{0.2599} & \textbf{0.4049} & \textbf{0.3165} & \textbf{0.5094} \\
    \bottomrule
  \end{tabular}}
  \vspace{2mm}
  \caption{\textbf{Zero-Shot 2D Semantic Segmentation Experiments on ScanNet20~\cite{dai2017scannet} (20 classes),  ScanNet200~\cite{dai2017scannet} (200 classes) and \scannetpp~\cite{yeshwanth2023scannet++}.} All methods are evaluated on a 10-scene validation set.}
  \label{tab:2D_eval}
  \vspace{-7mm}
\end{table}

\begin{table}[htbp]
  \centering
  \setlength{\tabcolsep}{16pt}% column sep.
  \setlength{\extrarowheight}{0.1pt}
  \small
  \scalebox{0.8}{
  \begin{tabular}{@{}l
                  cc   % 10 scenes: f‑mIoU | f‑mAcc
                  cc 
                  @{}}
    \toprule
    \multirow{2.4}{*}{\textbf{Method}}
      & \multicolumn{2}{c}{\textbf{HoliCity (10 scenes)}}
      & \multicolumn{2}{c}{\textbf{Holicity (328 scenes)}} 

      \\
    \cmidrule{2-3} \cmidrule{4-5}
      & f–mIoU & f–mAcc & f–mIoU & f–mAcc \\
    \midrule
    \rowcolor{tblgray}
    \multicolumn{5}{l}{\textit{Per‑Scene Optimization Methods}} \\
    LangSplat~\cite{Qin2024LangSplat}              & 0.0918      & 0.1950      & -      & -     \\
    OpenGaussian~\cite{Wu2024OpenGaussians}           & 0.1853      & 0.2567      & -     & -      \\
    Feature‑3DGS~\cite{zhou2024feature3DGS}           & 0.2479      & 0.5646     & -      & -   \\
    FMGS~\cite{Zuo2024FMGS}                   & 0.1774 & 0.2513 & -  & -    \\
    GOI~\cite{QU2024GOI}                    & 0.1126      & 0.3773     & 0.1587      & 0.3772     \\
    \rowcolor{tblgray}
    \multicolumn{5}{l}{\textit{Per‑Scene Optimization‑Free Methods}} \\
    Gradient-Weighted 3DGS~\cite{joseph2024GWfeatureBP}    & 0.2484      & 0.4808      & 0.2762      & 0.4187      \\
    LUDVIG~\cite{marrie2024ludvig}                 & 0.1593 & 0.2296 & 0.1843 & 0.2775 \\
    OccamLGS~\cite{cheng2024occam}                & 0.2255 & 0.4754 & 0.2234 & 0.5056 \\
    \textcolor{gray}{SceneSplat~\cite{li2025SceneSplat7k} (Pseudo Label)} & \textcolor{gray}{0.2464} & \textcolor{gray}{0.4780} & \textcolor{gray}{0.2522} & \textcolor{gray}{0.5159} \\
    \rowcolor{tblgray}
    \multicolumn{5}{l}{\textit{Generalizable Method}} \\
    SceneSplat~\cite{li2025SceneSplat7k}             & \textbf{0.3081} & \textbf{0.5647} & \textbf{0.2880} & \textbf{0.6045} \\
    \bottomrule
  \end{tabular}
  }
  \vspace{2mm}
  \caption{\textbf{Zero-Shot 3D Outdoor Segmentation On \holi~\cite{zhou2021holicity} Dataset.} All methods are evaluated on mini-validation sets,with the full validation set evaluated only for selected methods due to runtime limitations (\cref{tab:benchmark_overview}).}
  \label{tab:3D_eval_holicity}
  \vspace{-3mm}
\end{table}

\begin{figure}[ht]
  \centering
\includegraphics[width=0.9\linewidth]{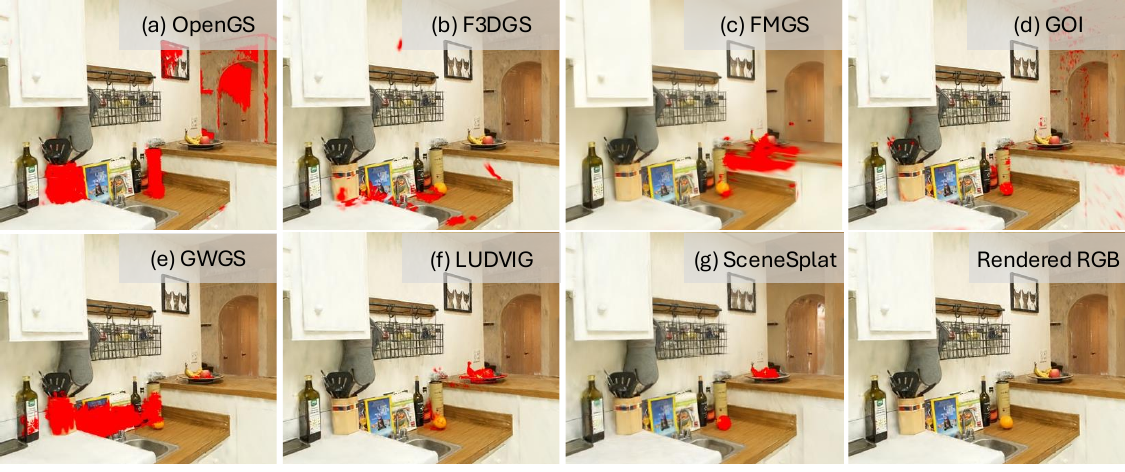}
  \caption{\textbf{Text-Based Scene Query.} Given the prompt "These are fruits" to different LGS methods, the queried parts are highlighted in red.}
  \label{fig:vis_text_query}
\end{figure}

\begin{table}[htbp]
  \centering
   \setlength{\tabcolsep}{16pt}% column sep.
  \setlength{\extrarowheight}{0.1pt}
  \small
  \scalebox{0.8}{
  \begin{tabular}{@{}l cc| cc @{}}
    \toprule
    \multirow{2.4}{*}{\textbf{Method}} &
      \multicolumn{2}{c|}{\textbf{ScanNet (10 scenes)}} &
      \multicolumn{2}{c}{\textbf{ScanNet++ (10 scenes)}} \\
    \cmidrule(lr){2-3}\cmidrule(lr){4-5}
    & Accuracy (BBox) & Accuracy (Seg) &
      Accuracy (BBox) & Accuracy (Seg) \\
    \midrule
    \rowcolor{tblgray}
    \multicolumn{5}{l}{\textit{Per-Scene Optimization Methods}} \\
    LangSplat~\cite{Qin2024LangSplat} & 0.0663 & 0.0415 & 0.0655 & 0.0337 \\
    OpenGaussian~\cite{Wu2024OpenGaussians} & 0.4565 & 0.3483 & 0.5138 & 0.3394 \\
    Feature-3DGS~\cite{zhou2024feature3DGS} & 0.5861 & 0.4759 & 0.5046 & 0.2783 \\
    FMGS~\cite{Zuo2024FMGS} & 0.6082 & 0.4639 & \textcolor{gray}{0.5494} & 0.3303 \\
    GOI~\cite{QU2024GOI} & 0.5269 & 0.4414 & 0.5107 & \textcolor{gray}{0.3568} \\
    \rowcolor{tblgray}
    \multicolumn{5}{l}{\textit{Per-Scene Optimization-Free Methods}} \\
    Gradient-Weighted 3DGS~\cite{joseph2024GWfeatureBP} & 0.5486 & 0.4669 & 0.4322 & 0.2661 \\
    LUDVIG~\cite{marrie2024ludvig} & \textcolor{gray}{0.6750} & \textcolor{gray}{0.5481} & 0.4801 & 0.2875 \\
    OccamLGS~\cite{cheng2024occam} & 0.2074 & 0.1599 & 0.1896 & 0.1182 \\
    \rowcolor{tblgray}
    \multicolumn{5}{l}{\textit{Generalizable Method}} \\
    \textbf{SceneSplat}~\cite{li2025SceneSplat7k} & \textbf{0.7701} & \textbf{0.6928} & \textbf{0.7339} & \textbf{0.5464} \\
    \bottomrule
  \end{tabular}%
  }
  \vspace{2mm}
  \caption{\textbf{3D Object Localization Experiments on \scannet~\cite{dai2017scannet} and \scannetpp~\cite{yeshwanth2023scannet++}}. Accuracy is reported with bounding box–based and segmentation–based evaluation.}
  \label{tab:3d_localization}
  \vspace{-3mm}
\end{table}

\begin{table*}[!t]
  \centering
  \small
  \resizebox{\linewidth}{!}{%
    \setlength{\tabcolsep}{2pt}%
    \setlength{\extrarowheight}{0.1pt}
    \begin{tabular}{@{}
                  l            % training source
                  l            % # scenes
                  cc           % ScanNet++ 
                  cc           % ScanNet200 
                  cc           % Matterport3D
                  cc           % HoliCity (out-of-domain)
                  @{}}
    \toprule
    \multirow{2}{*}{\textbf{Training Source}} &
    \multirow{2}{*}{\textbf{\#Scene}} &
      \multicolumn{2}{C{3cm}}{\textbf{ScanNet++}} &
      \multicolumn{2}{C{3cm}}{\textbf{ScanNet200}} &
      \multicolumn{2}{C{3cm}}{\textbf{Matterport3D}} &
      \multicolumn{2}{C{3cm}}{\textbf{HoliCity}} \\
    \cmidrule{3-4}\cmidrule{5-6}\cmidrule{7-8}\cmidrule{9-10}
      & & f–mIoU & f–mAcc & f–mIoU & f–mAcc & f–mIoU & f–mAcc & mIoU & mAcc \\
    \midrule

    \rowcolor{tblgray}
    \multicolumn{10}{l}{\textit{Indoor Training-Data Scaling}} \\[0.1em]
    ScanNet++ (v1)                       & 280  & 0.1683 & 0.3230 & 0.1081 & 0.1650 & 0.1010 & 0.2108 & 0.1888 & 0.3003 \\
    ScanNet++ (v2)                       & 906 & 0.2383 & 0.4324 & 0.1180 & 0.1920 & 0.1491  & 0.2756  & 0.2339 & 0.3524 \\
    ScanNet++ (v2) + ScanNet             & 2107 & 0.2779 & \textbf{0.5055} & 0.1427 & 0.3434 & 0.2289 & 0.3803 & 0.2633 & 0.3504 \\
    ScanNet++ (v2) + ScanNet + M3D       & 3503 & \textbf{0.2836} & 0.4992 & \textbf{0.1648} & \textbf{0.3573} & \textbf{0.3384} & \textbf{0.5745} & 0.2026 & 0.3600 \\
    Matterport3D only                    &  1396  &  —   &  —   & —      & —      & 0.3244 & 0.5354 & 0.2240   &   0.3889    \\

    \rowcolor{tblgray}
    \multicolumn{10}{l}{\textit{Outdoor-Domain Training}} \\[0.1em]
    HoliCity only                        & 3000 & —      & —      & —      & —      & —      & —      & \textbf{0.2880} & \textbf{0.6045} \\
    \bottomrule
    \end{tabular}%
  }
    \vspace{1mm}
    \caption{\textbf{Impact of Training-Data Scaling on Indoor Benchmarks and Cross-Domain Generalization to HoliCity~\cite{zhou2021holicity}.} Results show that more training data consistently improves indoor performance. Furthermore, models trained on indoor data only surprisingly transfer to outdoor scenes.}
    \label{tab:scenesplat_scaling}
    \vspace{-1mm}
\end{table*}

\subsection{Key Insights}
\label{key_insights}
The primary experimental results from our \benchmark benchmark across all selected methods are presented in~\cref{tab:3D_eval_scannet,tab:3D_eval_scannetpp_matterport,tab:2D_eval} for indoor scenes and in~\cref{tab:3D_eval_holicity} for outdoor scenes. The corresponding runtime statistics for individual methods can be found in~\cref{tab:benchmark_overview}. SceneSplat -- representing the generalizable paradigm -- stands out as the clear winner in terms of both performance and efficiency (see \cref{fig:vis_sem_seg} for qualitative examples).  Interestingly, in 75\% of the experiments, it even outperforms its own pseudo labels used for pretraining (SceneSplat (Pseudo Label)). This result highlights its ability to generalize by leveraging large-scale data priors and learning to predict meaningful vision-language features, effectively mitigating the noise inherent in weak supervision. In addition, \cref{tab:2D_eval} shows that the performance on the 2D metric is aligned with the zero-shot performance measured with the 3D metric, with a generalizable method leading the performance. As illustrated in~\cref{fig:vis_text_query}, visualizations of text query results using enhanced NeRFView tools~\cite{nerfstudio,yang2024spectrally,li2025SceneSplat7k} further demonstrate that the generalizable method outperforms the others. Moreover, unlike other methods that rely on extensive feature extraction, the generalizable approach requires only 3DGS as input at inference time and predicts per-primitive vision-language features in a single forward pass.

Among the per-scene methods, optimization-free approaches clearly outperform optimization-based ones, both in terms of segmentation accuracy, as shown in~\cref{tab:3D_eval_scannet,tab:3D_eval_scannetpp_matterport,tab:3D_eval_holicity,tab:2D_eval}, and in runtime efficiency, as reported in~\cref{tab:benchmark_overview}. This is likely due to the fact that the objective function in optimization-based methods is typically designed to perform as well as possible on the training views, which may not be ideal to novel viewpoints. Taking this into account, along with the strong performance of the generalizable approach, we conclude that per-scene optimization is not necessary for effective vision-language reasoning in 3DGS. However, within both the optimization-based and optimization-free groups, no single method consistently dominates across all datasets, indicating that performance remains sensitive to dataset-specific characteristics.

A key requirement for reliable benchmarking is having a sufficient number of scenes and appropriately challenging evaluation settings. As shown in~\cref{tab:3D_eval_scannet,tab:3D_eval_scannetpp_matterport,tab:3D_eval_holicity}, the performance of various methods varies noticeably when scaling from a small subset of 10 scenes to the full validation set, highlighting the importance of the benchmark scale in assessing performance. Furthermore, ~\cref{tab:3D_eval_scannet} demonstrates a clear drop in accuracy when the task complexity increases from 20 to 200 semantic classes, emphasizing the need for demanding benchmarks to reveal the limitations of competing approaches.

To further diversify the open-vocabulary evaluation, we include a 3D object localization task that assesses how well each method retrieves objects in 3D space using its constructed 3DGS field.
Following the protocol of LangSplat~\cite{Qin2024LangSplat}, we directly select the point with the highest relevance score for each query.
Our evaluation is conducted on 2,895 objects from \scannet and \scannetpp validation scenes, which is over 12× more objects compared to LERF~\cite{kerr2023lerf}, offering a broader and more realistic assessment. 
This evaluation is presented in~\cref{tab:3d_localization} and shows similar conclusion as previously discussed evaluations. The generalizable paradigm shows dominant performance, followed by the per-scene optimization-free methods, which outperform per-scene optimization method group.
% \textcolor{yellow}{The results in ~\cref{tab:3d_localization} reveal a clear performance separation across different LGS methods.
% }

% \begin{itemize}
% \item 
%     Several disadvantages related to the used 2D metrics. 1.multi-view inconsistency, high accuracy in a set of 2D views does not necessarily indicate the correctness of the 3D segmentation. 2. Evaluation close to the views on which they were overfitted on.
% \item
%  2D and 3D metrics are complimentary and lead to a more comprehensive benchmark for LGS.
% \end{itemize}}

% \textcolor{yellow}{Q2:Do we want to mention noisy 3D annotations and evaluation complexities? There are still some space.}

\begin{figure}[hbtp]
  \centering
\includegraphics[width=0.9\linewidth]{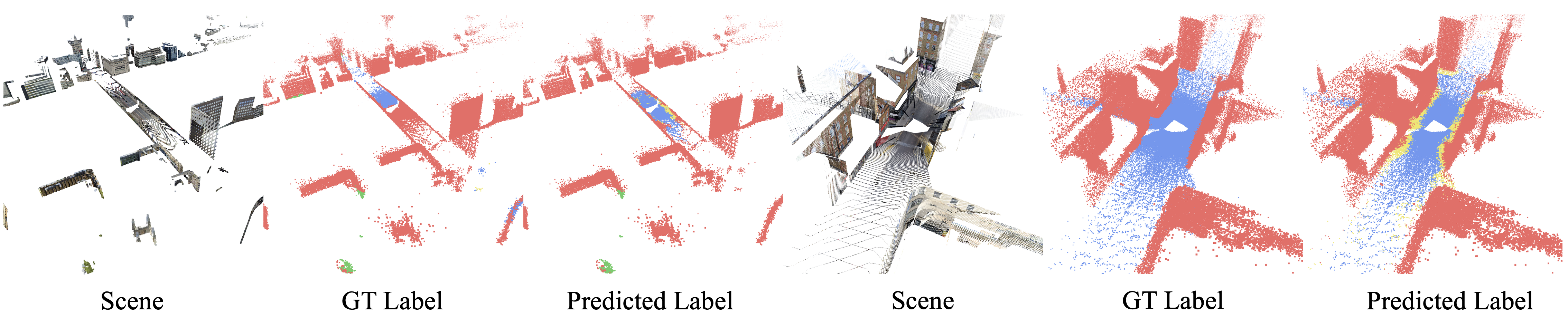}
  \caption{\textbf{Zero-Shot Predictions of Indoor-Trained SceneSplat on Outdoor Scenes.} The results highlight the cross-domain capability. Color palette denotes \textcolor[rgb]{0.95, 0.4, 0.4}{buildings}, \textcolor[rgb]{0.4, 0.6, 0.95}{roads}, \textcolor[rgb]{0.95, 0.9, 0.4}{terrains}, and \textcolor[rgb]{0.4, 0.8, 0.4}{trees}.}
  \label{fig:holicity_vis}
\end{figure}

\cref{tab:scenesplat_scaling} presents our scaling study, which offers two key takeaways. First, training-data scaling consistently improves performance on indoor benchmarks: expanding the training set from 280 to 3503 scenes lifts ScanNet++ f-mIoU from 0.168 $\rightarrow$ 0.284 and ScanNet200 from 0.108 $\rightarrow$ 0.165. A similar effect appears when the model jointly trained on three indoor datasets outperforms that trained on Matterport3D alone.
Second—and more surprisingly—an indoor-only model transfers to outdoor scenes, but domain-specific data is still required to close the gap. Without ever seeing outdoor data, the largest indoor model reaches 0.263 mIoU on HoliCity, an improvement of 7.4\% over the 280-scene baseline. This suggests that large-scale indoor training induces representations that partially generalize across the indoor–outdoor boundary, see~\cref{fig:holicity_vis} for zero-shot examples. Nevertheless, a model trained only on HoliCity still leads (0.288 mIoU), indicating that cross-domain capability ultimately needs outdoor supervision.
These findings underscore the value of our large-scale indoor–outdoor 3DGS dataset as a powerful training resource.

%merge 4. benchmark and 5.experments
% 4 Benchmark and Experiments
%4.1 Problem Formulation
% ·open-volcabulary understanding task
% input & output
% ·baseline & 3D evaluation
% ·dataset and metric we used
%4.2 Experimental Settings
%4.3 Experimental results(on 3 indoor datasets and 1 outdoor dataset)
%4.4 Dicussion and Key findings

%-----------------------%
\section{Conclusion and Limitations}
\label{sec:conclusion}
%-----------------------%

This work introduces the first comprehensive benchmark for 3DGS-based scene understanding methods, along with a large-scale 3DGS dataset of diverse indoor and outdoor scenes. Our evaluation demonstrates that generalizable approach consistently outperforms per-scene methods, establishing a new paradigm for scalable scene understanding through pretrained models. Despite these contributions, limitations remain. Our dataset scale, while substantial, could be further expanded. The outdoor benchmark is constrained by limited semantic classes, and not all 3DGS scenes include precomputed vision-language embeddings. We leave these as important directions for future work and hope our benchmark and dataset will be helpful to advance language-grounded 3DGS scene understanding.
%-----------------------%

% \input{secs/author_guideline}

%%%%%%%%%%% Acknowledgement %%%%%%%%%%
\begin{ack}
% We thank ...
% This work was partially supported by ...
Yue Li is financially supported by TomTom, the University of Amsterdam and the allowance of Top consortia for Knowledge and Innovation (TKIs) from the Netherlands Ministry of Economic Affairs and Climate Policy. This work used Dutch national e-infrastructure with the support of the SURF Cooperative under grant no. NWO-2024.035.
This work was partially supported by INSAIT, Sofia University “St. Kliment Ohridski” (Partially funded by the Ministry of Education and Science of Bulgaria’s support for INSAIT as part of the Bulgarian National Roadmap for Research Infrastructure), 
EU Horizon projects ELIAS (No.101120237), 
and ELLIOT (No.101214398).
Some computational resources were provided by the Google Cloud Platform (GCP). 
\end{ack}

%%%%%%%%%%% Reference %%%%%%%%%%
\clearpage
\bibliographystyle{plain}
\bibliography{main}

@String(CVPR= {IEEE Conf. Comput. Vis. Pattern Recog.})

@String(ECCV= {Eur. Conf. Comput. Vis.})

@String(ICLR = {Int. Conf. Learn. Represent.})

@String(CVPR  = {CVPR})

@String(ECCV  = {ECCV})

@String(ICLR  = {ICLR})

@InProceedings{Qin2024LangSplat,
    author    = {Qin, Minghan and Li, Wanhua and Zhou, Jiawei and Wang, Haoqian and Pfister, Hanspeter},
    title     = {LangSplat: 3D Language Gaussian Splatting},
    booktitle = {Proceedings of the IEEE/CVF Conference on Computer Vision and Pattern Recognition (CVPR)},
    month     = {June},
    year      = {2024},
    pages     = {20051-20060}
}

@inproceedings{Wu2024OpenGaussians,
 author = {Wu, Yanmin and Meng, Jiarui and Li, Haijie and Wu, Chenming and Shi, Yahao and Cheng, Xinhua and Zhao, Chen and Feng, Haocheng and Ding, Errui and Wang, Jingdong and Zhang, Jian},
 booktitle = {Advances in Neural Information Processing Systems},
 title = {OpenGaussian: Towards Point-Level 3D Gaussian-based Open Vocabulary Understanding},
 volume = {37},
 year = {2024}
}

@article{zuo2024FMGS,
  title={Fmgs: Foundation model embedded 3d gaussian splatting for holistic 3d scene understanding},
  author={Zuo, Xingxing and Samangouei, Pouya and Zhou, Yunwen and Di, Yan and Li, Mingyang},
  journal={arXiv preprint arXiv:2401.01970},
  year={2024}
}

@article{shi4568335city,
  title={City-Scale Mapping System with Three-Layer Sampling And Panoptic Representation},
  author={Shi, Yongliang and Yang, Runyi and Wu, Zirui and Li, Pengfei and Liu, Caiyun and Zhao, Hao and Zhou, Guyue},
  year = {2023},
  journal={Available at SSRN 4568335}
}

@inproceedings{zhou2024feature3DGS,
    title={Feature 3dgs: Supercharging 3d gaussian splatting to enable distilled feature fields},
    author={Zhou, Shijie and Chang, Haoran and Jiang, Sicheng and Fan, Zhiwen and Zhu, Zehao and Xu, Dejia and Chari, Pradyumna and You, Suya and Wang, Zhangyang and Kadambi, Achuta},
    booktitle={Proceedings of the IEEE/CVF Conference on Computer Vision and Pattern Recognition},
    pages={21676--21685},
    year={2024}
}

@article{QU2024GOI,
    title={GOI: Find 3D Gaussians of Interest with an Optimizable Open-vocabulary Semantic-space Hyperplane},
    author={Qu, Yansong and Dai, Shaohui and Li, Xinyang and Lin, Jianghang and Cao, Liujuan and Zhang, Shengchuan and Ji, Rongrong},
    journal={arXiv preprint arXiv:2405.17596},
    year={2024}
}

@misc{joseph2024GWfeatureBP,
    title={Gradient-Weighted Feature Back-Projection: A Fast Alternative to Feature Distillation in 3D Gaussian Splatting}, 
    author={Joji Joseph and Bharadwaj Amrutur and Shalabh Bhatnagar},
    year={2024},
    eprint={2411.15193},
    archivePrefix={arXiv},
    primaryClass={cs.CV},
    url={https://arxiv.org/abs/2411.15193}, 
}

@article{marrie2024ludvig,
    title={LUDVIG: Learning-free uplifting of 2d visual features to Gaussian splatting scenes},
    author={Marrie, Juliette and Menegaux, Romain and Arbel, Michael and Larlus, Diane and Mairal, Julien},
    journal={arXiv preprint arXiv:2410.14462},
    year={2024}
}

@misc{li2025SceneSplat7k,
      title={SceneSplat: Gaussian Splatting-based Scene Understanding with Vision-Language Pretraining}, 
      author={Yue Li and Qi Ma and Runyi Yang and Huapeng Li and Mengjiao Ma and Bin Ren and Nikola Popovic and Nicu Sebe and Ender Konukoglu and Theo Gevers and Luc Van Gool and Martin R. Oswald and Danda Pani Paudel},
      year={2025},
      eprint={2503.18052},
      archivePrefix={arXiv},
      primaryClass={cs.CV},
      url={https://arxiv.org/abs/2503.18052}, 
}

@inproceedings{Ye2024GSgroupping,
    title={Gaussian Grouping: Segment and Edit Anything in 3D Scenes},
    author={Ye, Mingqiao and Danelljan, Martin and Yu, Fisher and Ke, Lei},
    booktitle={ECCV},
    year={2024}
}

@InProceedings{Shi2024LEG,
    author    = {Shi, Jin-Chuan and Wang, Miao and Duan, Hao-Bin and Guan, Shao-Hua},
    title     = {Language Embedded 3D Gaussians for Open-Vocabulary Scene Understanding},
    booktitle = {Proceedings of the IEEE/CVF Conference on Computer Vision and Pattern Recognition (CVPR)},
    month     = {June},
    year      = {2024},
    pages     = {5333-5343}
}

@inproceedings{jiang2024open,
  title={Open-vocabulary 3d semantic segmentation with foundation models},
  author={Jiang, Li and Shi, Shaoshuai and Schiele, Bernt},
  booktitle={Proceedings of the IEEE/CVF Conference on Computer Vision and Pattern Recognition},
  pages={21284--21294},
  year={2024}
}

@Article{kerbl3Dgaussians,
      author       = {Kerbl, Bernhard and Kopanas, Georgios and Leimk{\"u}hler, Thomas and Drettakis, George},
      title        = {3D Gaussian Splatting for Real-Time Radiance Field Rendering},
      journal      = {ACM Transactions on Graphics},
      number       = {4},
      volume       = {42},
      month        = {July},
      year         = {2023},
      url          = {https://repo-sam.inria.fr/fungraph/3d-gaussian-splatting/}
}

@inproceedings{ding2022pla,
  title={Pla: Language-driven open-vocabulary 3d scene understanding},
  author={Ding, Runyu and Yang, Jihan and Xue, Chuhui and Zhang, Wenqing and Bai, Song and Qi, Xiaojuan},
  booktitle={Proceedings of the IEEE/CVF conference on computer vision and pattern recognition},
  pages={7010--7019},
  year={2023}
}

@article{yang2024spectrally,
  title={Spectrally pruned gaussian fields with neural compensation},
  author={Yang, Runyi and Zhu, Zhenxin and Jiang, Zhou and Ye, Baijun and Chen, Xiaoxue and Zhang, Yifei and Chen, Yuantao and Zhao, Jian and Zhao, Hao},
  journal={arXiv preprint arXiv:2405.00676},
  year={2024}
}

@inproceedings{yang2024regionplc,
  title={Regionplc: Regional point-language contrastive learning for open-world 3d scene understanding},
  author={Yang, Jihan and Ding, Runyu and Deng, Weipeng and Wang, Zhe and Qi, Xiaojuan},
  booktitle={Proceedings of the IEEE/CVF Conference on Computer Vision and Pattern Recognition},
  pages={19823--19832},
  year={2024}
}

@inproceedings{nerfstudio,
	title        = {Nerfstudio: A Modular Framework for Neural Radiance Field Development},
	author       = {
		Tancik, Matthew and Weber, Ethan and Ng, Evonne and Li, Ruilong and Yi, Brent
		and Kerr, Justin and Wang, Terrance and Kristoffersen, Alexander and Austin,
		Jake and Salahi, Kamyar and Ahuja, Abhik and McAllister, David and Kanazawa,
		Angjoo
	},
	year         = 2023,
	booktitle    = {ACM SIGGRAPH 2023 Conference Proceedings},
	series       = {SIGGRAPH '23}
}

@inproceedings{dai2017scannet,
  title={Scannet: Richly-annotated 3d reconstructions of indoor scenes},
  author={Dai, Angela and Chang, Angel X and Savva, Manolis and Halber, Maciej and Funkhouser, Thomas and Nie{\ss}ner, Matthias},
  booktitle={Proceedings of the IEEE conference on computer vision and pattern recognition},
  pages={5828--5839},
  year={2017}
}

@inproceedings{yeshwanth2023scannet++,
  title={Scannet++: A high-fidelity dataset of 3d indoor scenes},
  author={Yeshwanth, Chandan and Liu, Yueh-Cheng and Nie{\ss}ner, Matthias and Dai, Angela},
  booktitle={Proceedings of the IEEE/CVF International Conference on Computer Vision},
  pages={12--22},
  year={2023}
}

@article{chang2017matterport3d,
  title={Matterport3d: Learning from rgb-d data in indoor environments},
  author={Chang, Angel and Dai, Angela and Funkhouser, Thomas and Halber, Maciej and Niessner, Matthias and Savva, Manolis and Song, Shuran and Zeng, Andy and Zhang, Yinda},
  journal={arXiv preprint arXiv:1709.06158},
  year={2017}
}

@misc{zhou2021holicity,
      title={HoliCity: A City-Scale Data Platform for Learning Holistic 3D Structures}, 
      author={Yichao Zhou and Jingwei Huang and Xili Dai and Shichen Liu and Linjie Luo and Zhili Chen and Yi Ma},
      year={2021},
      eprint={2008.03286},
      archivePrefix={arXiv},
      primaryClass={cs.CV},
      url={https://arxiv.org/abs/2008.03286}, 
}

@inproceedings{charatan23pixelsplat,
      title={pixelSplat: 3D Gaussian Splats from Image Pairs for Scalable Generalizable 3D Reconstruction},
      author={David Charatan and Sizhe Li and Andrea Tagliasacchi and Vincent Sitzmann},
      year={2024},
      booktitle={CVPR},
}

@article{chen2024mvsplat,
    title   = {MVSplat: Efficient 3D Gaussian Splatting from Sparse Multi-View Images},
    author  = {Chen, Yuedong and Xu, Haofei and Zheng, Chuanxia and Zhuang, Bohan and Pollefeys, Marc and Geiger, Andreas and Cham, Tat-Jen and Cai, Jianfei},
    journal = {arXiv preprint arXiv:2403.14627},
    year    = {2024},
}

@article{gslrm2024,
    author={Zhang, Kai and Bi, Sai and Tan, Hao and Xiangli, Yuanbo and Zhao, Nanxuan 
      and Sunkavalli, Kalyan and Xu, Zexiang},
    title     = {GS-LRM: Large Reconstruction Model for 3D Gaussian Splatting},
    journal   = {European Conference on Computer Vision},
    year      = {2024},
}

@article{ziwen2024llrm,
  title={Long-LRM: Long-sequence Large Reconstruction Model for Wide-coverage Gaussian Splats},
  author={Ziwen, Chen and Tan, Hao and Zhang, Kai and Bi, Sai and Luan, Fujun and Hong, Yicong and Fuxin, Li and Xu, Zexiang},
  journal={arXiv preprint 2410.12781},
  year={2024}
}

@inproceedings{roessle2024l3dg,
      title={L3DG: Latent 3D Gaussian Diffusion}, 
      author={Roessle, Barbara and M{\"u}ller, Norman and Porzi, Lorenzo and Bul{\`o}, Samuel Rota and Kontschieder, Peter and Dai, Angela and Nie{\ss}ner, Matthias},
      booktitle={SIGGRAPH Asia 2024 Conference Papers},
      month={December},
      year={2024}
}

@article{ye2025gsplat,
  title={gsplat: An open-source library for Gaussian splatting},
  author={Vickie Ye and Ruilong Li and Justin Kerr and Matias Turkulainen and Brent Yi and Zhuoyang Pan and Otto Seiskari and Jianbo Ye and Jeffrey Hu and Matthew Tancik and Angjoo Kanazawa},
  journal={Journal of Machine Learning Research},
  volume={26},
  number={34},
  pages={1--17},
  year={2025},
  url={https://arxiv.org/abs/2409.06765}, 
}

@InProceedings{He_2016_CVPR,
author = {He, Kaiming and Zhang, Xiangyu and Ren, Shaoqing and Sun, Jian},
title = {Deep Residual Learning for Image Recognition},
booktitle = {Proceedings of the IEEE Conference on Computer Vision and Pattern Recognition (CVPR)},
month = {June},
year = {2016}
}

@article{dosovitskiy2020vit,
  title={An Image is Worth 16x16 Words: Transformers for Image Recognition at Scale},
  author={Dosovitskiy, Alexey and Beyer, Lucas and Kolesnikov, Alexander and Weissenborn, Dirk and Zhai, Xiaohua and Unterthiner, Thomas and  Dehghani, Mostafa and Minderer, Matthias and Heigold, Georg and Gelly, Sylvain and Uszkoreit, Jakob and Houlsby, Neil},
  journal={ICLR},
  year={2021}
}

@misc{oquab2023dinov2,
  title={DINOv2: Learning Robust Visual Features without Supervision},
  author={Oquab, Maxime and Darcet, Timothée and Moutakanni, Theo and Vo, Huy V. and Szafraniec, Marc and Khalidov, Vasil and Fernandez, Pierre and Haziza, Daniel and Massa, Francisco and El-Nouby, Alaaeldin and Howes, Russell and Huang, Po-Yao and Xu, Hu and Sharma, Vasu and Li, Shang-Wen and Galuba, Wojciech and Rabbat, Mike and Assran, Mido and Ballas, Nicolas and Synnaeve, Gabriel and Misra, Ishan and Jegou, Herve and Mairal, Julien and Labatut, Patrick and Joulin, Armand and Bojanowski, Piotr},
  journal={arXiv:2304.07193},
  year={2023}
}

@article{ravi2024sam2,
  title={SAM 2: Segment Anything in Images and Videos},
  author={Ravi, Nikhila and Gabeur, Valentin and Hu, Yuan-Ting and Hu, Ronghang and Ryali, Chaitanya and Ma, Tengyu and Khedr, Haitham and R{\"a}dle, Roman and Rolland, Chloe and Gustafson, Laura and Mintun, Eric and Pan, Junting and Alwala, Kalyan Vasudev and Carion, Nicolas and Wu, Chao-Yuan and Girshick, Ross and Doll{\'a}r, Piotr and Feichtenhofer, Christoph},
  journal={arXiv preprint arXiv:2408.00714},
  url={https://arxiv.org/abs/2408.00714},
  year={2024}
}

@inproceedings{lyu2024unibind,
  title={Unibind: Llm-augmented unified and balanced representation space to bind them all},
  author={Lyu, Yuanhuiyi and Zheng, Xu and Zhou, Jiazhou and Wang, Lin},
  booktitle={Proceedings of the IEEE/CVF Conference on Computer Vision and Pattern Recognition},
  pages={26752--26762},
  year={2024}
}

@article{lyu2025realrag,
  title={RealRAG: Retrieval-augmented Realistic Image Generation via Self-reflective Contrastive Learning},
  author={Lyu, Yuanhuiyi and Zheng, Xu and Jiang, Lutao and Yan, Yibo and Zou, Xin and Zhou, Huiyu and Zhang, Linfeng and Hu, Xuming},
  journal={arXiv preprint arXiv:2502.00848},
  year={2025}
}

@InProceedings{Radford2021CLIP,
  title = 	 {Learning Transferable Visual Models From Natural Language Supervision},
  author =       {Radford, Alec and Kim, Jong Wook and Hallacy, Chris and Ramesh, Aditya and Goh, Gabriel and Agarwal, Sandhini and Sastry, Girish and Askell, Amanda and Mishkin, Pamela and Clark, Jack and Krueger, Gretchen and Sutskever, Ilya},
  booktitle = 	 {Proceedings of the 38th International Conference on Machine Learning},
  pages = 	 {8748--8763},
  year = 	 {2021},
  volume = 	 {139},
  series = 	 {Proceedings of Machine Learning Research}
}

@inproceedings{liu2023llava,
author      = {Liu, Haotian and Li, Chunyuan and Wu, Qingyang and Lee, Yong Jae},
title       = {Visual Instruction Tuning},
booktitle   = {NeurIPS},
year        = {2023}
}

@article{xiang2024structured,
    title   = {Structured 3D Latents for Scalable and Versatile 3D Generation},
    author  = {Xiang, Jianfeng and Lv, Zelong and Xu, Sicheng and Deng, Yu and Wang, Ruicheng and Zhang, Bowen and Chen, Dong and Tong, Xin and Yang, Jiaolong},
    journal = {arXiv preprint arXiv:2412.01506},
    year    = {2024}
}

@article{zhong2025omnisam,
  title={OmniSAM: Omnidirectional Segment Anything Model for UDA in Panoramic Semantic Segmentation},
  author={Zhong, Ding and Zheng, Xu and Liao, Chenfei and Lyu, Yuanhuiyi and Chen, Jialei and Wu, Shengyang and Zhang, Linfeng and Hu, Xuming},
  journal={arXiv preprint arXiv:2503.07098},
  year={2025}
}

@inproceedings{wu2025sonata,
    title={Sonata: Self-Supervised Learning of Reliable Point Representations},
    author={Wu, Xiaoyang and DeTone, Daniel and Frost, Duncan and Shen, Tianwei and Xie, Chris and Yang, Nan and Engel, Jakob and Newcombe, Richard and Zhao, Hengshuang and Straub, Julian},
    booktitle={CVPR},
    year={2025}
}

@article{wu2023mars,
  author    = {Wu, Zirui and Liu, Tianyu and Luo, Liyi and Zhong, Zhide and Chen, Jianteng and Xiao, Hongmin and Hou, Chao and Lou, Haozhe and Chen, Yuantao and Yang, Runyi and Huang, Yuxin and Ye, Xiaoyu and Yan, Zike and Shi, Yongliang and Liao, Yiyi and Zhao, Hao},
  title     = {MARS: An Instance-aware, Modular and Realistic Simulator for Autonomous Driving},
  journal   = {CICAI},
  year      = {2023},
}

@inproceedings{mildenhall2020nerf,
 title={NeRF: Representing Scenes as Neural Radiance Fields for View Synthesis},
 author={Ben Mildenhall and Pratul P. Srinivasan and Matthew Tancik and Jonathan T. Barron and Ravi Ramamoorthi and Ren Ng},
 year={2020},
 booktitle={ECCV},
}

@inproceedings{sarkar2025crossover,
    author={Sayan Deb Sarkar and Ondrej Miksik and Marc Pollefeys and Daniel Barath and Iro Armeni},
    title={CrossOver: 3D Scene Cross-Modal Alignment}, 
    booktitle = {The IEEE Conference on Computer Vision and Pattern Recognition (CVPR)},
    year = {2025}
}

@INPROCEEDINGS{Newcombe2011KinectFusion,
  author={Newcombe, Richard A. and Izadi, Shahram and Hilliges, Otmar and Molyneaux, David and Kim, David and Davison, Andrew J. and Kohi, Pushmeet and Shotton, Jamie and Hodges, Steve and Fitzgibbon, Andrew},
  booktitle={2011 10th IEEE International Symposium on Mixed and Augmented Reality}, 
  title={KinectFusion: Real-time dense surface mapping and tracking}, 
  year={2011},
  pages={127-136},
  doi={10.1109/ISMAR.2011.6092378}}

@inproceedings{schoenberger2016sfm,
    author={Sch\"{o}nberger, Johannes Lutz and Frahm, Jan-Michael},
    title={Structure-from-Motion Revisited},
    booktitle={Conference on Computer Vision and Pattern Recognition (CVPR)},
    year={2016},
}

@INPROCEEDINGS{Park2019DeepSDF,
  author={Park, Jeong Joon and Florence, Peter and Straub, Julian and Newcombe, Richard and Lovegrove, Steven},
  booktitle={2019 IEEE/CVF Conference on Computer Vision and Pattern Recognition (CVPR)}, 
  title={DeepSDF: Learning Continuous Signed Distance Functions for Shape Representation}, 
  year={2019},
  pages={165-174},
  doi={10.1109/CVPR.2019.00025}}

@ARTICLE{Hirschmuller2008StereoDepth,
  author={Hirschmuller, Heiko},
  journal={IEEE Transactions on Pattern Analysis and Machine Intelligence}, 
  title={Stereo Processing by Semiglobal Matching and Mutual Information}, 
  year={2008},
  volume={30},
  number={2},
  pages={328-341},
  doi={10.1109/TPAMI.2007.1166}}

@inproceedings{Kazhdan2006, 
    author = {Kazhdan, Michael and Bolitho, Matthew and Hoppe, Hugues}, title = {Poisson surface reconstruction}, 
    year = {2006}, 
    isbn = {3905673363}, 
    booktitle = {Proceedings of the Fourth Eurographics Symposium on Geometry Processing}, 
    pages = {61–70}, 
    numpages = {10}, 
    series = {SGP '06} 
}

@inproceedings{ma2024shapesplat,
      title={ShapeSplat: A Large-scale Dataset of Gaussian Splats and Their Self-Supervised Pretraining}, 
      author={Qi Ma and Yue Li and Bin Ren and Nicu Sebe and Ender Konukoglu and Theo Gevers and Luc Van Gool and Danda Pani Paudel},
      booktitle={3DV},
      year={2024}
}

@inproceedings{liu24uco3d,
    Author = {Liu, Xingchen and Tayal, Piyush and Wang, Jianyuan
              and Zarzar, Jesus and Monnier, Tom and Tertikas, Konstantinos
              and Duan, Jiali and Toisoul, Antoine and Zhang, Jason Y.
              and Neverova, Natalia and Vedaldi, Andrea
              and Shapovalov, Roman and Novotny, David},
    Booktitle = {arXiv},
    Title = {UnCommon Objects in 3D},
    Year = {2024},
}

@INPROCEEDINGS{8099499,
  author={Charles, R. Qi and Su, Hao and Kaichun, Mo and Guibas, Leonidas J.},
  booktitle={2017 IEEE Conference on Computer Vision and Pattern Recognition (CVPR)}, 
  title={PointNet: Deep Learning on Point Sets for 3D Classification and Segmentation}, 
  year={2017},
  pages={77-85},
  doi={10.1109/CVPR.2017.16}}

@inproceedings{wu2024ptv3,
    title={Point Transformer V3: Simpler, Faster, Stronger},
    author={Wu, Xiaoyang and Jiang, Li and Wang, Peng-Shuai and Liu, Zhijian and Liu, Xihui and Qiao, Yu and Ouyang, Wanli and He, Tong and Zhao, Hengshuang},
    booktitle={CVPR},
    year={2024}
}

@inproceedings{rozenberszki2022language,
  title={Language-grounded indoor 3d semantic segmentation in the wild},
  author={Rozenberszki, David and Litany, Or and Dai, Angela},
  booktitle={European Conference on Computer Vision},
  pages={125--141},
  year={2022},
  organization={Springer}
}

@inproceedings{peng2023openscene,
  title={Openscene: 3d scene understanding with open vocabularies},
  author={Peng, Songyou and Genova, Kyle and Jiang, Chiyu and Tagliasacchi, Andrea and Pollefeys, Marc and Funkhouser, Thomas and others},
  booktitle={Proceedings of the IEEE/CVF conference on computer vision and pattern recognition},
  pages={815--824},
  year={2023}
}

@inproceedings{zhu2023pointclip,
  title={Pointclip v2: Prompting clip and gpt for powerful 3d open-world learning},
  author={Zhu, Xiangyang and Zhang, Renrui and He, Bowei and Guo, Ziyu and Zeng, Ziyao and Qin, Zipeng and Zhang, Shanghang and Gao, Peng},
  booktitle={Proceedings of the IEEE/CVF international conference on computer vision},
  pages={2639--2650},
  year={2023}
}

@article{baruch2021arkitscenes,
  title={Arkitscenes: A diverse real-world dataset for 3d indoor scene understanding using mobile rgb-d data},
  author={Baruch, Gilad and Chen, Zhuoyuan and Dehghan, Afshin and Dimry, Tal and Feigin, Yuri and Fu, Peter and Gebauer, Thomas and Joffe, Brandon and Kurz, Daniel and Schwartz, Arik and others},
  journal={arXiv preprint arXiv:2111.08897},
  year={2021}
}

@inproceedings{roberts2021hypersim,
  title={Hypersim: A photorealistic synthetic dataset for holistic indoor scene understanding},
  author={Roberts, Mike and Ramapuram, Jason and Ranjan, Anurag and Kumar, Atulit and Bautista, Miguel Angel and Paczan, Nathan and Webb, Russ and Susskind, Joshua M},
  booktitle={Proceedings of the IEEE/CVF international conference on computer vision},
  pages={10912--10922},
  year={2021}
}

@inproceedings{wald2019rio,
  title={Rio: 3d object instance re-localization in changing indoor environments},
  author={Wald, Johanna and Avetisyan, Armen and Navab, Nassir and Tombari, Federico and Nie{\ss}ner, Matthias},
  booktitle={Proceedings of the IEEE/CVF International Conference on Computer Vision},
  pages={7658--7667},
  year={2019}
}

@inproceedings{ren2024bringing,
  title={Bringing masked autoencoders explicit contrastive properties for point cloud self-supervised learning},
  author={Ren, Bin and Mei, Guofeng and Paudel, Danda Pani and Wang, Weijie and Li, Yawei and Liu, Mengyuan and Cucchiara, Rita and Van Gool, Luc and Sebe, Nicu},
  booktitle={Proceedings of the Asian Conference on Computer Vision},
  pages={2034--2052},
  year={2024}
}

@inproceedings{PointMAE,
  title={Masked autoencoders for point cloud self-supervised learning},
  author={Pang, Yatian and Wang, Wenxiao and Tay, Francis EH and Liu, Wei and Tian, Yonghong and Yuan, Li},
  booktitle={European conference on computer vision},
  pages={604--621},
  year={2022},
  organization={Springer}
}

@inproceedings{zhao2021point,
  title={Point transformer},
  author={Zhao, Hengshuang and Jiang, Li and Jia, Jiaya and Torr, Philip HS and Koltun, Vladlen},
  booktitle={Proceedings of the IEEE/CVF international conference on computer vision},
  pages={16259--16268},
  year={2021}
}

@article{wu2022point,
  title={Point transformer v2: Grouped vector attention and partition-based pooling},
  author={Wu, Xiaoyang and Lao, Yixing and Jiang, Li and Liu, Xihui and Zhao, Hengshuang},
  journal={Advances in Neural Information Processing Systems},
  volume={35},
  pages={33330--33342},
  year={2022}
}

@article{zhang2022dino,
  title={Dino: Detr with improved denoising anchor boxes for end-to-end object detection},
  author={Zhang, Hao and Li, Feng and Liu, Shilong and Zhang, Lei and Su, Hang and Zhu, Jun and Ni, Lionel M and Shum, Heung-Yeung},
  journal={arXiv preprint arXiv:2203.03605},
  year={2022}
}

@inproceedings{kirillov2023segment,
  title={Segment anything},
  author={Kirillov, Alexander and Mintun, Eric and Ravi, Nikhila and Mao, Hanzi and Rolland, Chloe and Gustafson, Laura and Xiao, Tete and Whitehead, Spencer and Berg, Alexander C and Lo, Wan-Yen and others},
  booktitle={Proceedings of the IEEE/CVF international conference on computer vision},
  pages={4015--4026},
  year={2023}
}

@article{ravi2024sam,
  title={Sam 2: Segment anything in images and videos},
  author={Ravi, Nikhila and Gabeur, Valentin and Hu, Yuan-Ting and Hu, Ronghang and Ryali, Chaitanya and Ma, Tengyu and Khedr, Haitham and R{\"a}dle, Roman and Rolland, Chloe and Gustafson, Laura and others},
  journal={arXiv preprint arXiv:2408.00714},
  year={2024}
}

@inproceedings{radford2021learning,
  title={Learning transferable visual models from natural language supervision},
  author={Radford, Alec and Kim, Jong Wook and Hallacy, Chris and Ramesh, Aditya and Goh, Gabriel and Agarwal, Sandhini and Sastry, Girish and Askell, Amanda and Mishkin, Pamela and Clark, Jack and others},
  booktitle={International conference on machine learning},
  pages={8748--8763},
  year={2021},
  organization={PmLR}
}

@inproceedings{zhai2023sigmoid,
  title={Sigmoid loss for language image pre-training},
  author={Zhai, Xiaohua and Mustafa, Basil and Kolesnikov, Alexander and Beyer, Lucas},
  booktitle={Proceedings of the IEEE/CVF international conference on computer vision},
  pages={11975--11986},
  year={2023}
}

@article{tschannen2025siglip,
  title={SigLIP 2: Multilingual Vision-Language Encoders with Improved Semantic Understanding, Localization, and Dense Features},
  author={Tschannen, Michael and Gritsenko, Alexey and Wang, Xiao and Naeem, Muhammad Ferjad and Alabdulmohsin, Ibrahim and Parthasarathy, Nikhil and Evans, Talfan and Beyer, Lucas and Xia, Ye and Mustafa, Basil and others},
  journal={arXiv preprint arXiv:2502.14786},
  year={2025}
}

@article{straub2019replica,
  title={The replica dataset: A digital replica of indoor spaces},
  author={Straub, Julian and Whelan, Thomas and Ma, Lingni and Chen, Yufan and Wijmans, Erik and Green, Simon and Engel, Jakob J and Mur-Artal, Raul and Ren, Carl and Verma, Shobhit and others},
  journal={arXiv preprint arXiv:1906.05797},
  year={2019}
}

@inproceedings{kerr2023lerf,
  title={Lerf: Language embedded radiance fields},
  author={Kerr, Justin and Kim, Chung Min and Goldberg, Ken and Kanazawa, Angjoo and Tancik, Matthew},
  booktitle={Proceedings of the IEEE/CVF International Conference on Computer Vision},
  pages={19729--19739},
  year={2023}
}

@article{cheng2024occam,
  title={Occam's LGS: A Simple Approach for Language Gaussian Splatting},
  author={Cheng, Jiahuan and Zaech, Jan-Nico and Van Gool, Luc and Paudel, Danda Pani},
  journal={arXiv preprint arXiv:2412.01807},
  year={2024}
}

@article{zheng2024gaussiangrasper,
  title={Gaussiangrasper: 3d language gaussian splatting for open-vocabulary robotic grasping},
  author={Zheng, Yuhang and Chen, Xiangyu and Zheng, Yupeng and Gu, Songen and Yang, Runyi and Jin, Bu and Li, Pengfei and Zhong, Chengliang and Wang, Zengmao and Liu, Lina and others},
  journal={IEEE Robotics and Automation Letters},
  year={2024},
  publisher={IEEE}
}

@article{guo2024semantic,
  title={Semantic gaussians: Open-vocabulary scene understanding with 3d gaussian splatting},
  author={Guo, Jun and Ma, Xiaojian and Fan, Yue and Liu, Huaping and Li, Qing},
  journal={arXiv preprint arXiv:2403.15624},
  year={2024}
}

@inproceedings{rashid2023language,
  title={Language embedded radiance fields for zero-shot task-oriented grasping},
  author={Rashid, Adam and Sharma, Satvik and Kim, Chung Min and Kerr, Justin and Chen, Lawrence Yunliang and Kanazawa, Angjoo and Goldberg, Ken},
  booktitle={7th Annual Conference on Robot Learning},
  year={2023}
}

@article{kerbl20233d,
  title={3d gaussian splatting for real-time radiance field rendering.},
  author={Kerbl, Bernhard and Kopanas, Georgios and Leimk{\"u}hler, Thomas and Drettakis, George},
  journal={ACM Trans. Graph.},
  volume={42},
  number={4},
  pages={139--1},
  year={2023}
}

@article{mildenhall2021nerf,
  title={Nerf: Representing scenes as neural radiance fields for view synthesis},
  author={Mildenhall, Ben and Srinivasan, Pratul P and Tancik, Matthew and Barron, Jonathan T and Ramamoorthi, Ravi and Ng, Ren},
  journal={Communications of the ACM},
  volume={65},
  number={1},
  pages={99--106},
  year={2021},
  publisher={ACM New York, NY, USA}
}

@article{lee2025mosaic3d,
  title={Mosaic3D: Foundation Dataset and Model for Open-Vocabulary 3D Segmentation},
  author={Lee, Junha and Park, Chunghyun and Choe, Jaesung and Wang, Yu-Chiang Frank and Kautz, Jan and Cho, Minsu and Choy, Chris},
  journal={arXiv preprint arXiv:2502.02548},
  year={2025}
}

@misc{scannet200,
      title={Language-Grounded Indoor 3D Semantic Segmentation in the Wild}, 
      author={David Rozenberszki and Or Litany and Angela Dai},
      year={2022},
      eprint={2204.07761},
      archivePrefix={arXiv},
      primaryClass={cs.CV},
      url={https://arxiv.org/abs/2204.07761}, 
}

@article{kheradmand20253d,
  title={3d gaussian splatting as markov chain monte carlo},
  author={Kheradmand, Shakiba and Rebain, Daniel and Sharma, Gopal and Sun, Weiwei and Tseng, Yang-Che and Isack, Hossam and Kar, Abhishek and Tagliasacchi, Andrea and Yi, Kwang Moo},
  journal={Advances in Neural Information Processing Systems},
  volume={37},
  pages={80965--80986},
  year={2025}
}

@misc{wei20253daffordsplatefficientaffordancereasoning,
      title={3DAffordSplat: Efficient Affordance Reasoning with 3D Gaussians}, 
      author={Zeming wei and Junyi Lin and Yang Liu and Weixing Chen and Jingzhou Luo and Guanbin Li and Liang Lin},
      year={2025},
      eprint={2504.11218},
      archivePrefix={arXiv},
      primaryClass={cs.CV},
      url={https://arxiv.org/abs/2504.11218}, 
}

@misc{chen2024splatformer,
    title = {SplatFormer: Point Transformer for Robust 3D Gaussian Splatting},
    author = {Chen, Yutong and Mihajlovic, Marko and Chen, Xiyi and Wang, Yiming and Prokudin, Sergey and Tang, Siyu},
    booktitle = {International Conference on Learning Representations (ICLR)},
    year = {2025}
}

@article{WITKIN198117,
  title = {Recovering surface shape and orientation from texture},
  journal = {Artificial Intelligence},
  volume = {17},
  number = {1},
  pages = {17-45},
  year = {1981},
  issn = {0004-3702},
  doi = {https://doi.org/10.1016/0004-3702(81)90019-9},
  author = {Andrew P. Witkin},
}

@book{10.5555/539405, 
    author = {Grimson, Eric L. W.}, 
    title = {From Images to Surfaces: A Computational Study of the Human Early Visual System}, 
    year = {1981}, 
    isbn = {0262070839}, 
    publisher = {MIT Press}, 
    address = {Cambridge, MA, USA} 
}

@INPROCEEDINGS{791235,
  author={Kutulakos, K.N. and Seitz, S.M.},
  booktitle={Proceedings of the Seventh IEEE International Conference on Computer Vision}, 
  title={A theory of shape by space carving}, 
  year={1999},
  volume={1},
  number={},
  pages={307-314 vol.1},
  doi={10.1109/ICCV.1999.791235}
}

@ARTICLE{5226635,
  author={Furukawa, Yasutaka and Ponce, Jean},
  journal={IEEE Transactions on Pattern Analysis and Machine Intelligence}, 
  title={Accurate, Dense, and Robust Multiview Stereopsis}, 
  year={2010},
  volume={32},
  number={8},
  pages={1362-1376},
  doi={10.1109/TPAMI.2009.161}
}

@misc{liu2024weaklysupervised3dopenvocabulary,
      title={Weakly Supervised 3D Open-vocabulary Segmentation}, 
      author={Kunhao Liu and Fangneng Zhan and Jiahui Zhang and Muyu Xu and Yingchen Yu and Abdulmotaleb El Saddik and Christian Theobalt and Eric Xing and Shijian Lu},
      year={2024},
      eprint={2305.14093},
      archivePrefix={arXiv},
      primaryClass={cs.CV},
      url={https://arxiv.org/abs/2305.14093}, 
}

@article{ma2025cityloc,
  title={CityLoc: 6DoF Pose Distributional Localization for Text Descriptions in Large-Scale Scenes with Gaussian Representation},
  author={Ma, Qi and Yang, Runyi and Ren, Bin and Sebe, Nicu and Konukoglu, Ender and Van Gool, Luc and Paudel, Danda Pani},
  journal={arXiv preprint arXiv:2501.08982},
  year={2025}
}

@misc{barron2021mipnerf,
      title={Mip-NeRF: A Multiscale Representation for Anti-Aliasing Neural Radiance Fields},
      author={Jonathan T. Barron and Ben Mildenhall and Matthew Tancik and Peter Hedman and Ricardo Martin-Brualla and Pratul P. Srinivasan},
      year={2021},
      eprint={2103.13415},
      archivePrefix={arXiv},
      primaryClass={cs.CV}
}

@inproceedings{ren2022neural,
  title={Neural volumetric object selection},
  author={Ren, Zhongzheng and Agarwala, Aseem and Russell, Bryan and Schwing, Alexander G and Wang, Oliver},
  booktitle={Proceedings of the IEEE/CVF Conference on Computer Vision and Pattern Recognition},
  pages={6133--6142},
  year={2022}
}

@inproceedings{spinnerf,
      title={{SPIn-NeRF}: Multiview Segmentation and Perceptual Inpainting with Neural Radiance Fields}, 
      author={Ashkan Mirzaei and Tristan Aumentado-Armstrong and Konstantinos G. Derpanis and Jonathan Kelly and Marcus A. Brubaker and Igor Gilitschenski and Alex Levinshtein},
      year={2023},
      booktitle={CVPR},
}

@inproceedings{ling2024dl3dv,
  title={Dl3dv-10k: A large-scale scene dataset for deep learning-based 3d vision},
  author={Ling, Lu and Sheng, Yichen and Tu, Zhi and Zhao, Wentian and Xin, Cheng and Wan, Kun and Yu, Lantao and Guo, Qianyu and Yu, Zixun and Lu, Yawen and others},
  booktitle={Proceedings of the IEEE/CVF Conference on Computer Vision and Pattern Recognition},
  pages={22160--22169},
  year={2024}
}

@inproceedings{avetisyan2024scenescript,
  title={Scenescript: Reconstructing scenes with an autoregressive structured language model},
  author={Avetisyan, Armen and Xie, Christopher and Howard-Jenkins, Henry and Yang, Tsun-Yi and Aroudj, Samir and Patra, Suvam and Zhang, Fuyang and Frost, Duncan and Holland, Luke and Orme, Campbell and others},
  booktitle={European Conference on Computer Vision},
  pages={247--263},
  year={2024},
  organization={Springer}
}

@article{halacheva2025gaussianvlm,
  title={GaussianVLM: Scene-centric 3D Vision-Language Models using Language-aligned Gaussian Splats for Embodied Reasoning and Beyond},
  author={Halacheva, Anna-Maria and Zaech, Jan-Nico and Wang, Xi and Paudel, Danda Pani and Van Gool, Luc},
  journal={arXiv preprint arXiv:2507.00886},
  year={2025}
}

% !!! Important! MUST !!!
%%%%%%%%%%% Checklist %%%%%%%%%%
% \clearpage
% \input{secs/checklist}

%%%%%%%%%%% Appendix %%%%%%%%%%
% \clearpage
% \appendix

% \begin{center}
% \LARGE \textbf{SceneSplat++: A Large Dataset and Comprehensive\\
% Benchmark for Language Gaussian Splatting}\\[0.6em]
% \LARGE \textit{Supplementary Material}
% \end{center}

% \input{secs/Appendix}

\end{document}